\documentclass[10pt,journal,compsoc]{IEEEtran}

\usepackage{graphicx}
\usepackage{amsmath}
\usepackage{amssymb}
\usepackage{booktabs}

\usepackage{times}
\usepackage{epsfig}
\usepackage{graphicx}
\usepackage{amsmath}
\usepackage{amssymb}
\usepackage{algorithm}
\usepackage{algpseudocode}
\usepackage{threeparttable}
\usepackage{multirow}
\usepackage{booktabs}
\usepackage{caption}
\usepackage{mathrsfs}
\usepackage{bbding}
\usepackage[pagebackref,breaklinks,colorlinks]{hyperref}

\begin{document}
	%
	\title{Variational Cross-Graph Reasoning and Adaptive Structured Semantics Learning for Compositional Temporal Grounding}
	%
	%
	%
	%
	
    \author{
	Juncheng Li,
	Siliang Tang,
	Linchao Zhu,
	Wenqiao Zhang,
	Yi Yang,~\IEEEmembership{Senior Member,~IEEE,}\\
	Tat-Seng Chua,
	Fei Wu,~\IEEEmembership{Senior Member,~IEEE,}
	Yueting Zhuang,~\IEEEmembership{Senior Member,~IEEE}
	\IEEEcompsocitemizethanks{
		\IEEEcompsocthanksitem J. Li, S. Tang, L. Zhu, Y. Yang, F. Wu, and Y. Zhuang are with  Zhejiang University, Hangzhou, China. E-mail: \{junchengli, siliang, zhulinchao, yangyics, wufei, yzhuang\}@zju.edu.cn.
		\IEEEcompsocthanksitem W. Zhang and T. Chua are with the National University of Singapore, Singapore. E-mail: \{wenqiao, dcscts\}@nus.edu.sg.
	}
}
	%
	%

	\markboth{Journal of \LaTeX\ Class Files,~Vol.~xxx, No.~xxx, September~2022}%
	{Shell \MakeLowercase{\textit{et al.}}: Bare Demo of IEEEtran.cls for Computer Society Journals}
	%



	\IEEEtitleabstractindextext{
		
		\begin{abstract}
			Temporal grounding is the task of  locating a specific segment from an untrimmed video according to a query sentence. This task has achieved significant momentum in the computer vision community as it enables activity grounding beyond pre-defined activity classes by utilizing the semantic diversity of natural language descriptions. The semantic diversity is rooted in the principle of compositionality in linguistics, where novel semantics can be systematically described by combining known words in novel ways~(\textbf{compositional generalization}). However, existing temporal grounding datasets are not carefully designed to evaluate the compositional generalizability. To systematically benchmark the compositional generalizability of temporal grounding models, we introduce a new Compositional Temporal Grounding task and construct two new dataset splits, \textsl{i.e.}, Charades-CG and ActivityNet-CG. When evaluating the state-of-the-art methods on our new dataset splits, we empirically find that they fail to generalize to queries with novel combinations of seen words. We argue that the inherent structured semantics inside the videos and language is the crucial factor to achieve compositional generalization. Based on this insight, we propose a variational cross-graph reasoning framework that explicitly decomposes video and language into hierarchical semantic graphs, respectively, and learns fine-grained semantic correspondence between the two graphs. Furthermore, we introduce a novel adaptive structured semantics learning approach to derive the structure-informed and domain-generalizable graph representations, which facilitate the fine-grained semantic correspondence reasoning between the two graphs. Extensive experiments validate the superior compositional generalizability of our approach.
		\end{abstract}
		
		
	}

	\maketitle

	\IEEEdisplaynontitleabstractindextext

	%
	\IEEEpeerreviewmaketitle

	\IEEEraisesectionheading{\section{Introduction}\label{sec:introduction}}

	%
	%
	%
	%
	\IEEEPARstart{U}{nderstanding} rich and diverse activities in videos is an essential and fundamental goal of video understanding. While there have been substantial works in activity recognition~\cite{carreira2017quo, feichtenhofer2016convolutional} and localization~\cite{ma2016learning, singh2016multi}, one major limitation of these works is that they are restricted to pre-defined action classes, thus suffering from scaling to various complex activities. A natural solution to this problem is to utilize the systematic compositionality~\cite{fodor1988connectionism, chomsky2009syntactic, montague1970universal} of human language, which enables us to form novel compositions by combining known words in novel ways to describe unseen activities~(\textsl{i.e.,} \textbf{compositional generalization}). Therefore, a new task, namely temporal grounding in videos~\cite{gao2017tall, krishna2017dense}, has recently attracted significant attention from researchers. Formally, given an untrimmed video and a language query, the task aims to locate video segments that semantically correspond to the given query sentence.
	
	\begin{figure}[!t]
		\centering
		\includegraphics[width=\linewidth]{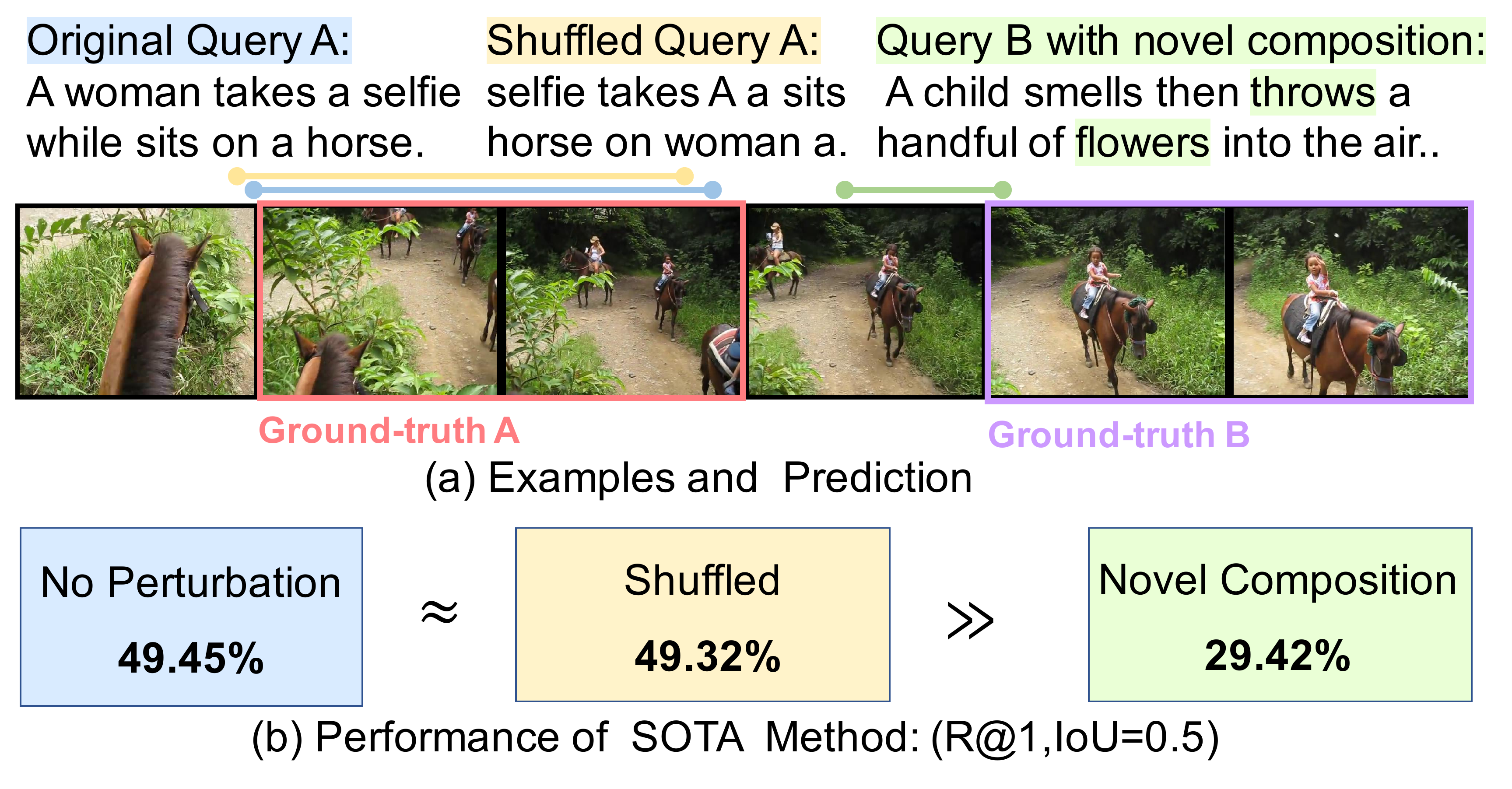}
		\caption{(a) On the top, we show three examples of two queries. (b) On the bottom, we report comparisons on Charades-CG with metric R@1, IoU@0.5. The left blue box represents the original model. The middle yellow box represents the model with shuffled queries as input. The right green box represents the performance on the queries that contain novel compositions.}
		\label{demo}
	\end{figure}
	
	Although the compositional generalization is a key property of human language that allows temporal grounding beyond pre-defined categories, existing temporal grounding datasets are not specifically designed to evaluate this ability. There is a large number of overlapping compositions~(\textsl{e.g.,} \textsl{verb-noun pair, adjective-noun pair, etc}) between the training and testing splits of existing datasets. Our statistical results show that only 1.37\% and 5.19\% of testing sentences contain novel compositions in the widely-used Charades-STA~\cite{gao2017tall} and ActivityNet Captions~\cite{krishna2017dense} datasets, respectively. To systematically benchmark the compositional generalizability~(CG) of temporal grounding models, we introduce a new task, \textbf{Compositional Temporal Grounding}. The proposed compositional temporal grounding task aims to test whether the model can generalize to sentences that contain novel compositions of seen words. We construct two re-organized datasets, \textsl{i.e.,} \textbf{Charades-CG} and \textbf{ActivityNet-CG}. Our dataset split protocols enable us to measure whether a model can generalize to novel compositions, of which the individual components have been observed during training while the combination is novel. Taking  \textsl{Query B} in Figure~\ref{demo} (a) as an example, \textsl{``throws"} and \textsl{``flowers"} are both observed separately in the training split, but the composition \textsl{``throws flowers''} has never been seen before. 
	
	Based on our newly constructed datasets, we evaluate modern state-of-the-art~(SOTA) temporal grounding models and empirically observe that SOTA models fail to achieve compositional generalization, though they have achieved promising progress on the standard temporal grounding task. We find that their performance drops dramatically~(Figure \ref{demo} (b), left vs. right) on the query sentences that contain novel compositions. The results indicate that SOTA models may not well generalize to novel compositions. Furthermore, as word order is a crucial factor for the compositionality of language, we analyze the word order sensitivity of SOTA models to gain more intuitive insight. Specifically, we randomly shuffled queries in advance and then train and evaluate the models using the shuffled sentences. Surprisingly, we find that evaluated models are insensitive to the word order, even though the complete semantics of original sentences are destroyed by permuting the word order~(Figure \ref{demo} (b), left vs. middle). These observations confirm with recent studies~\cite{otani2020uncovering, yuan2021closer}, which suggest that current models are heavily driven by superficial correlations. This pushes us to rethink the solution of temporal grounding.
	
	When we systematically analyze the SOTA methods, we find that previous temporal grounding models generally ignore the structured semantics in video and language, which, however, is crucial for compositional reasoning. These methods~\cite{mun2020local, zhang2020learning, zhang2020span, gao2017tall} mainly encode both sentences and video segments into unstructured global representations, respectively, and then employ advanced cross-modal fusion modules to obtain the final prediction. Such global representations fail to explicitly model video structure and language compositions. Take the novel composition of ``throws flowers''  in Figure \ref{demo} (a) as an example. If the model captures the correspondence between the semantics of the two words and relevant visual semantics in the video~(\textsl{i.e.,} \textsl{the action ``throw'' and the object ``flower'' in the video }), and understands the relationship between the components of the composition, the model can easily localize the novel composition in the video by composing the corresponding visual semantics of the two words. 
	
	
	Motivated by this insight, we propose a novel \textbf{V}ar\textbf{I}ational cro\textbf{S}s-graph re\textbf{A}soning~(\textbf{VISA}) framework for compositional temporal grounding.  Our VISA framework explicitly models the semantic structures of video and language, and establishes the fine-grained correspondence between them, thus achieving joint compositional reasoning. Specifically, we first introduce a \textbf{hierarchical semantic graph} that explicitly decomposes both video and language into three semantic hierarchies~(\textsl{i.e.,} \textsl{global events, local actions, and atomic objects}). 
	The initialized  hierarchical semantic graphs are then visual- and semantic- contextualized beyond separate semantic labels. The contextualized hierarchical semantic graph serves as the unified structured representations for both video and language, which tightly couple multi-granularity semantics between the two modalities. Second, we propose a \textbf{variational cross-graph correspondence reasoning} to infer the fine-grained semantic correspondence between the two graphs, which provides explicit guidance for grounding compositional semantics into videos.
	
	As the structured semantics represented by our hierarchical semantic graph is crucial for compositional reasoning, we further propose a new \textbf{Adaptive Structured Semantics Learning~(ASSL)} approach, which promotes the VISA model to capture the structure-informed and domain-generalizable semantics from video and language while simultaneously addressing the two potential limitations of VISA framework, which are, \textbf{1)~Semantic node noise:} the semantic nodes of video hierarchical semantic graph are initialized by detection models, of which the predictions might be noisy. Such undesirable noise from detection models might degrade the generalizability of the model. \textbf{2)~Cross-graph semantic shift:} the semantic nodes of two graphs reside in different semantic spaces, where the semantic labels of video graph nodes are defined by the predicted categories of detection models while the semantic labels of language graph nodes are drawn from verbs and noun phrases in sentences. For instance, the predicted action \textsl{``running''} from detection models might be referred as \textsl{``rush'' or ``dash''} in query sentences. Such semantic label shift makes it challenging to learn the semantic correspondence between the two graphs.
	
	Specifically, ASSL consists of two learning objectives: structure-informed semantics modeling and cross-graph semantic adaptation. For \textbf{structure-informed semantics modeling}, we randomly mask out some nodes~(\textsl{i.e., action nodes, object nodes}) in the video hierarchical semantic graph, and the goal is to predict these masked semantic nodes based on their surrounding nodes. These surrounding nodes serve as the semantic context that guides the model to encode relevant visual context from videos so that the model can complete the masked semantic nodes. This objective enforces the model to capture structured semantics from videos by jointly reasoning over the relationship with neighboring semantic nodes in the graph. To improve learning under noisy semantic nodes, we propose a structure-informed self-training strategy, which prevents the model from being over-penalized for predicting other reasonable results. Specifically, we keep a moving-averaged target network to compute the relationship between each masked node and other nodes and take the relationship distribution as pseudo targets. Instead of only learning to predict one-hot vectors corresponding to the detection labels, we use the more informative pseudo targets as additional supervision that encourages the model to investigate the semantic structure information and learn structure-informed representations. For \textbf{cross-graph semantic adaptation}, we introduce a semantic adaptation layer inserted between the hierarchical semantic graph and variational cross-graph correspondence reasoning. Guided by a novel semantic-adaptive margin loss, the semantic adaptation layer derives domain generalizable representations, which facilitates the fine-grained semantic correspondence reasoning between the two graphs. 
	
	This paper is a substantial and systematic extension of our previous CVPR version~\cite{li2022compositional} with the following improvements: 
	
	\begin{itemize}
		\item We propose the structure-informed semantics modeling that mitigates the semantic node noise and learns structure-informed representations in a self-training manner.
		
		\item We introduce the cross-graph semantic adaptation to address the cross-graph semantic shift problem and derive domain generalizable representations.
		
		
		\item We achieve new state-of-the-art performance on the proposed benchmarks and conduct more experiments to further verify the compositional generalizability and superiority of our approach, including more evaluation, ablations, complexity analysis, and visualizations.
		
	\end{itemize}

	\begin{figure*}[!t]
		\centering
		\includegraphics[width=\textwidth]{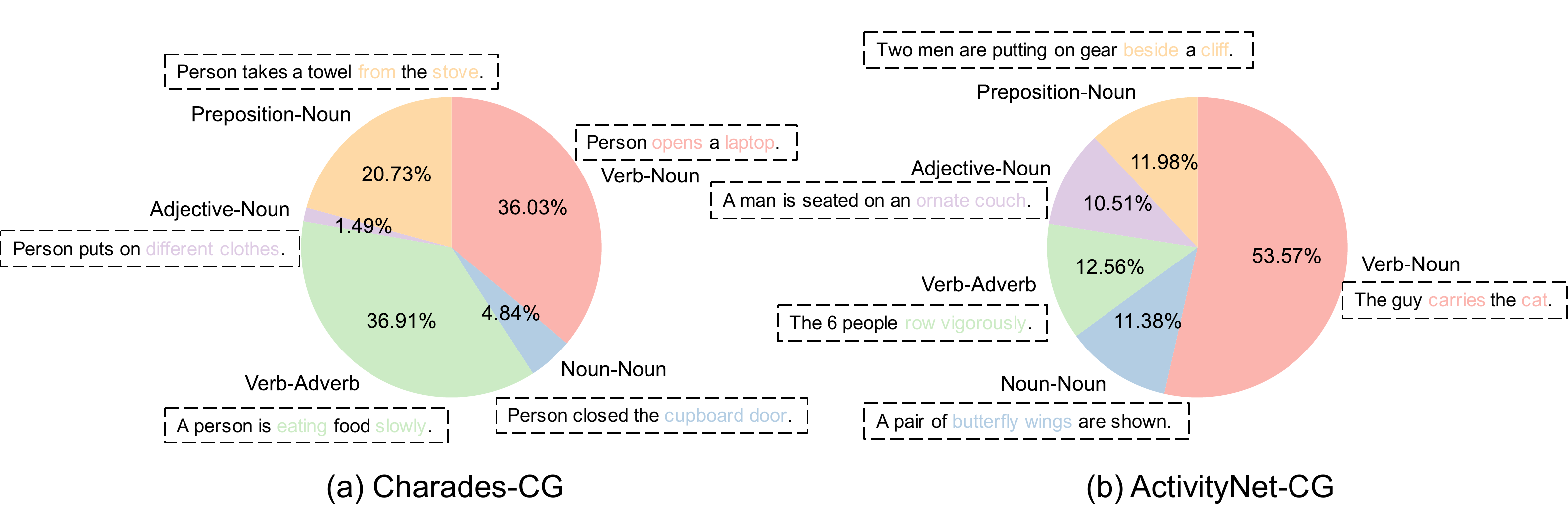}
		\caption{The distribution of the composition types. Text inside dashed boxes are query examples for each composition type.}
		\label{novel_type}
	\end{figure*}
	
	\section{Related Work}
	\noindent
	\textbf{Temporal Grounding.} Temporal grounding in videos via language is a recently proposed task~\cite{gao2017tall, krishna2017dense}. Existing supervised methods can be categorized into four groups. 1)~Proposal-based methods~\cite{gao2017tall, zhang2020learning, zhang2019man, chen2018temporally, yuan2019semantic} first extract candidate proposals by temporal sliding windows and then match the query sentence with them by multi-modality fusion. To further model the temporal dependencies among proposals, Zhang \textsl{et al.}~\cite{zhang2020learning} construct a 2D temporal adjacent map to generate multi-scale proposals and encode the adjacent temporal relation by Temporal Adjacent Network (2D-TAN). 
	2)~Proposal-free methods~\cite{mun2020local, zhang2020span, yuan2019find, zeng2020dense} directly predict the temporal boundaries of target segments without pre-defining proposals. Mun \textsl{et al.}~\cite{mun2020local} propose a local-global video-text interaction model that directly regresses the time interval. Zhang~\textsl{et al.}~\cite{zhang2020span} formulate temporal grounding as a span-based question answering task by treating the input video as text passage and propose a video span localizing network~(VSLNet) to predict the answer span. 3)~The RL-based methods~\cite{wu2020tree, wang2019language, he2019read} treat the problem as a sequential decision-making process and use Reinforcement Learning~(RL) algorithms to sequentially regulate the temporal boundaries of target segments. 4)~Since temporal boundary annotations are time-consuming, some works~\cite{duan2018weakly, gao2019wslln, mithun2019weakly, zhang2020counterfactual, song2020weakly, lin2020weakly} extend the problem to weakly-supervised setting by using multi-instance learning or language reconstruction as objective functions. However, it is unclear where the promising progress comes from. In this paper, we diagnose two representative SOTA methods and observe that they fail to capture the linguistic structure of language queries. Further, we systematically evaluate the compositional generalizability of current SOTA methods and find significant performance degradation. 
	
	
	\noindent
	\textbf{Compositional Generalization.} Recently, compositional generalization has received increasing attention because of its robustness and sample efficiency. To evaluate the compositional generalization, Lake~\textsl{et al.}~\cite{lake2018generalization} propose the SCAN benchmark, which requires translating instructions generated by a phrase-structure grammar to action sequences. The SCAN benchmark is split such that the testing set contains unseen compositions in the training set. The following works have proposed several techniques to improve SCAN, including data augmentation~\cite{andreas2019good}, meta-learning~\cite{lake2019compositional, nye2020learning}, and architectural design~\cite{gordon2019permutation, li2019compositional}. Some recent works also explore compositional generalization on other applications, including image captioning~\cite{nikolaus2019compositional}, visual question answering~\cite{grunde2021agqa, johnson2017clevr, bahdanau2019closure, hudson2018compositional}, and action recognition~\cite{materzynska2020something, zhukov2019cross, wray2019fine}. In this paper, we re-organize the two widely explored datasets and systematically study the compositional generalization on temporal grounding. 
	
	\noindent
	\textbf{Concept-Based Video Modeling.} The concept-based methods~\cite{dong2021dual, lai2014video, markatopoulou2017query, habibian2016video2vec} mainly utilize multiple pre-trained detection models to extract concepts~(\textsl{e.g.,} objects, scenes) on frames/segments and aggregate them to obtain concept-based video representations. An attractive property of concept-based representations is their good interpretability. However, these methods just regard the detection results as individual concept labels and simply aggregate them to form the video representation. In contrast, we learn a hierarchical semantic graph over these concepts to obtain fine-grained contextualized representations beyond the coarse semantic labels. Recently, Chen~\textsl{et al.}~\cite{chen2020fine} propose to use aligned text to guide video encoding into hierarchical levels by end-to-end learning. Our work differs as the nodes of video semantic graph is initialized by several pre-trained detection models, which is of better interpretability, and we formulate the graph matching into variational inference framework, where we model cross-graph semantic correspondence as latent variables.

	\begin{table*}[!htb]
		\caption{The detailed statistics of  the newly proposed Charades-CG and Activity-CG datasets.}
		\label{t1}
		\centering
		\begin{threeparttable}
			\begin{tabular}{ llcccc}
				\hline
				Dataset &Split  &Videos  &Average Video Length   &Queries &Average Query Length\\		
				
				\hline    
				\multirow{4}*{Charades-CG} 
				&Training                              &3555      &30.58s     &8281             &5.93\\
				&Novel-Composition              &2480      &30.70s       &3442             &6.86 \\
				&Novel-Word                         &588        &31.26s     &703               &7.24\\
				&Test-Trivial                          &1689       &30.82s      &3096           &5.96\\
				
				\hline
				\multirow{4}*{ActivityNet-CG}  
				&Training                            &9659          &116.94s     &36724         &13.33\\
				&Novel-Composition           &4202          &121.12s     &12028          &14.78 \\
				&Novel-Word                      &2011           &124.35s    &3944           &14.61\\
				&Test-Trivial                        &4775          &119.60s        &15712          &11.31 \\
				
				\hline
			\end{tabular}
		\end{threeparttable}
	\end{table*}

	\section{Compositional Temporal Grounding}

	\subsection{Problem Formulation}
	To systematically benchmark the compositional generalizability of existing temporal grounding methods, we introduce a new task, \textbf{Compositional Temporal Grounding}. Our compositional temporal grounding task targets at evaluating how well a model can generalize to query sentences that contain novel compositions or novel words. We re-organize two prevailing datasets Charades-STA~\cite{gao2017tall} and ActivityNet Captions~\cite{krishna2017dense}, named \textbf{Charades-CG} and \textbf{ActivityNet-CG}, respectively. Specifically, we define two new testing splits: Novel-Composition and Novel-Word. Each sentence in the novel-composition split contains one type of novel composition. We define the composition as novel composition if its constituents are both observed during training but their combination is novel. For instance, we might know the word ``pulling'' and ``horse'' by observing compositions ``pulling rope'' and ``lead horse'' in the query sentences of training split, respectively. However, in the novel-composition testing split, the model is asked to understand the composition ``pulling horse'', of which the individual words have been known during training while the combination is novel. Each sentence in the novel-word splits contains a novel word, which aims to test whether a model can infer the potential semantics of the unseen word based on the other learned composition components appearing in the context.

	\subsection{Dataset Re-splitting}
	For each dataset, we first combine all instances in the original training set and testing set, and remove the instances that can be easily predicted solely based on videos. We then re-split each dataset into four sets: training, novel-composition, novel-word, and test-trivial. The test-trivial set is similar to the original testing set, where most of the compositions are seen during training. Concretely, We use AllenNLP~\cite{gardner2018allennlp} to lemmatize and label all nouns, adjectives, verbs, adverbs, prepositions in language queries. Based on dependency parsing results, we define 5 types of compositions: \textsl{verb-noun, adjective-noun, noun-noun, verb-adverb}, and \textsl{preposition-noun}. For each type of composition, we construct a statistical table, where the row indexes are all possible first components of the composition and the column indexes are all possible second components of the composition. Taking verb-noun as an example, the element in row $i$ and column $j$ corresponds to the composition that consists of the $i-th$ verb and the $j-th$ noun in the dataset. For each table, we first sample an element from each row and each column, and then add all queries that contain the sampled compositions to the training set, which ensures that all components of compositions can be observed in the training set. Next, for each type of composition, we sample compositions from tables and add the corresponding queries into the novel-composition split. Meanwhile, we sample some words as new words and add the queries that contain the new words into the novel-word split. Since each video is associated with multiple text queries, if one query is selected into the training set, we will add other queries of the same video into the training set. If one query is selected into the novel-composition or novel-word split, we will add the remaining queries of the same video into the test-trivial set. Thus, we make sure that there is no video overlap between the training and testing sets. 
	
	Table \ref{t1} summarizes detailed statistics, and the distribution of the composition types and their corresponding examples are illustrated in Figure \ref{novel_type}. Note that adjective-noun phrases are rare in original Charades-STA dataset, and most of them are some high frequency phrases, so the proportion of novel adjective-noun compositions is relatively small in our Novel-Composition set. We provide more details on dataset splitting protocol and statistics in supplementary materials. 
%
%
%
%
%
	\begin{figure*}[!t]
		\centering
		\includegraphics[width=\textwidth]{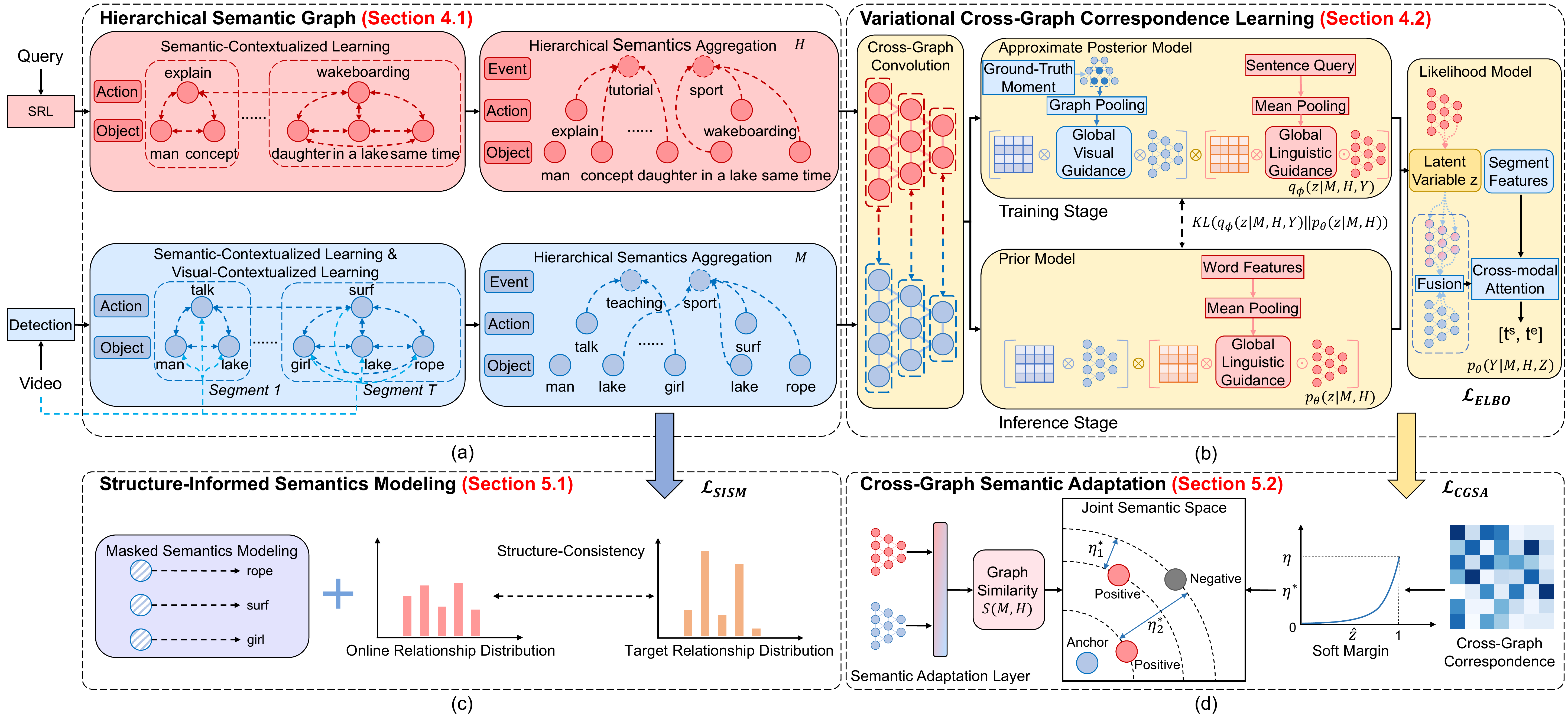}
		\caption{Overview of our variational cross-graph reasoning (VISA) and adaptive structured semantics learning (ASSL) framework. (a) Given a video and a language query, we decompose each of them into three semantic hierarchies and learn contextualized graph representations for them. (b) Then we infer fine-grained semantic correspondence between the two graphs. (c, d) The structure-informed semantics modeling and cross-graph semantic adaptation are further introduced to encourage structure-informed and domain-generalizable graph representations. Finally, we predict time intervals based on the learned graphs and their semantic correspondence. }
		\label{overview}
	\end{figure*}
	
	\section{Variational Cross-Graph Reasoning}\label{s4}
	As illustrated in Figure \ref{overview}, the variational cross-graph reasoning~(VISA) framework consists of two components: a \textbf{hierarchical semantic graph}~(Section~\ref{s4.1}) and a \textbf{variational cross-graph correspondence reasoning}~(Section~\ref{s4.1}). Given an untrimmed video $V$ and a query sentence $Q$, the hierarchical semantic graph first decomposes each of them into three semantic hierarchies~(\textsl{i.e.,} \textsl{global events, local actions, and atomic objects}). Then, the variational cross-graph correspondence reasoning establishes fine-grained semantic correspondence between two graphs. Finally, based on the fine-grained semantic correspondence between video and sentence, our VISA infers the target moment that semantically corresponds to the given query.
	
	\subsection{Hierarchical Semantic Graph}\label{s4.1}
	\vspace{-0.1cm}
	Language query sentences typically narrate several semantic events~\cite{li2021adaptive}, which can be further parsed to central predicates and their corresponding arguments. Similarly, videos naturally record some relevant events in our lives, which involves a variety of actions and each action relates to multiple objects. Therefore, language and videos are both inherently hierarchical in nature. Based on this observation, we decompose both of the input video and query into three semantic hierarchies, which correspond to global events, local actions, and atomic objects, respectively. Such a hierarchical semantic graph serves as a unified structured representation for modeling fine-grained semantic correspondence between videos and language queries. 
	

	\noindent
	\textbf{Graph Initialization.} For an untrimmed video $V$, we first divide it into a sequence of segments with a fixed length and adopt a pre-trained 3D CNN to extract the features : $\{V_t\}_{t=1}^T$, where $V_t=\{f_i^t\}_{i=1}^K$ and $f_i^t$ denotes the C3D features of frame $i$ in segment $t$. Then, we extract objects and actions from each segment by off-the-shelf object detection and action recognition models, where each segment contains $N_1$ object nodes $\{\overline{s}^{o}_{t, i}\}_{i=1}^{N_1} \in \mathbb{R}^{d \times N_1}$ and $N_2$ action nodes $\{\overline{s}^{a}_{t, i}\}_{i=1}^{N_2} \in \mathbb{R}^{d \times N_2}$. We initialize both object and action nodes by the sum of the GloVe~\cite{pennington2014glove} vectors of each word in the object labels and action labels. Finally, all the object nodes $\overline{S}^{o}$ and action nodes $\overline{S}^{a}$ across segments constitute the first and second hierarchies of the video hierarchical semantic graph.
	
	For query $Q$, we use semantic role labeling~(SRL) to parse the query into multiple semantic structures. Each semantic structure consists of a central predicate~(\textsl{i.e., verb}) and some corresponding arguments~(\textsl{i.e., noun phrases including prep, adj, and adv}). The predicates are considered as action nodes denoted by $\{\overline{c}^{a}_{i}\}_{i=1}^{L_2} \in \mathbb{R}^{d \times L_2}$, and the arguments are considered as object nodes denoted by $\{\overline{c}^{o}_{i, j}\}_{j=1}^{L_1} \in \mathbb{R}^{d \times L_1}$. If a word serves as multiple arguments for different predicates, we duplicate it for each predicate. Similarly, we initialize them using GloVe vectors. Finally, all the object nodes $\overline{C}^o$ and action nodes $\overline{C}^a$ constitute the first and second hierarchies of the language semantic graph.
	

	\noindent
	\textbf{Semantic-Contextualized Learning.} As the high-level semantic abstractions of video context, events involve complicated interactions between different semantic concepts. For example, the query ``the camel stands up and walks off with the family riding on its back'' is composed of objects~(\emph{camel}, \emph{the family}), actions~(\emph{stands up, walks off, and riding on}), and the underlying relations among them such as the spatial relation~(\emph{on its back}), the temporal relation~(\emph{stands up and walks off}), and the agentive relation~(\emph{riding on}). Thus, for comprehensive understanding of video events, we present semantic-contextualized learning to model the complicated interaction between the semantic nodes and learn rich contextual information beyond the coarse labels from detection models. Notably, semantic contextual information is crucial for resolving semantic ambiguity of individual semantic nodes as the detected actions and objects might have dramatic variations in appearance.
	
	Concretely, we define three types of undirect edges: \emph{action-action}, \emph{action-object}, and \emph{object-object}. For the video hierarchical semantic graph, the object nodes in the same segment are connected by the \emph{object-object} edges, the action and object nodes in the same segment are connected by the \emph{action-object} edges. For the language hierarchical semantic graph, the object nodes in the same semantic structure are connected by the \emph{object-object} edges, the action and object nodes in the same semantic structure are connected by the \emph{action-object} edges. All the action nodes of two graphs are connected by the \emph{action-action} edges, respectively. Afterward, we perform relation-aware graph convolution on video hierarchical semantic graph. For a semantic node $\overline{s}_i \in \{\overline{S}^a, \overline{S}^o\}$, we calculate the adjacency correlation for each edge type $r$ as:
	
	\begin{equation}
	\tilde{\alpha}_{ij}^r = (W^r \overline{s}_i)^T\cdot(W^r \overline{s}_j), \quad \alpha_{ij}^r  = \frac{exp(\tilde{\alpha}_{ij}^r)}{\sum_{j \in \mathcal{N}_i^r} exp(\tilde{\alpha}_{ij}^r)}
	\end{equation}
	
	\noindent
	where $\mathcal{N}_i^r$ is the neighborhood nodes of $s_i$ on edge type $r$ and $W_r$ is the relation-specific projection matrix. Then, we refine $s_i$ using the neighboring nodes of all edge types as: 
	
	\begin{equation}
	\hat{s}_i = \sum_{r \in R} \sum_{j \in \mathcal{N}_i^r} \alpha_{ij}^r \cdot (U^r \overline{s}_j)
	\end{equation}
	
	\noindent
	where $R$ represents three types of edges and $U_r$ is another transformation matrix. We obtain $\hat{s}_i$ after the first relation-aware graph convolution layer. To model multi-order relations, we perform multiple layers of relation-aware graph convolution and learn final semantic-contextualized node features $S=\{s_i\}_{i=1}^{N_v} \in \mathbb{R}^{N_v \times d}$, where $N_v$ is the total number of action and object nodes. In the same manner, we can obtain semantic-contextualized node features $C=\{c_i\}_{i=1}^{N_s} \in \mathbb{R}^{N_s \times d}$ of language semantic graph.
	
	\noindent
	\textbf{Visual-Contextualized Learning.} We further present visual-contextualized learning to enable the video semantic graph to gather relevant visual context from videos. Specifically, for a node $s_i$, let $V_i = \{f_j^i\}_{j=1}^K$ denotes the corresponding segment, and $f_j^i$ is the frame feature~(following, we omit the superscript i for simplicity). We first compute the visual filter for each frame $f_j$ in the segment and obtain the filtered visual feature as:
	
	\begin{equation}
	g^i_j = \sigma(W^g[s_i; \overline{f}; f_j] + b_g), \quad f_j^{'} = f_j \odot g^i_j
	\end{equation}
	
	\noindent
	where $\odot$ denotes the Hadamard product, and $\overline{f}$ is obtained by performing \verb|Avg-Pooling| over $V_i$. Then, we perform \verb|Max-Pooling| across the filtered frame features to get the semantic-relevant visual context as $F_i = MaxPool(f_1^{'}, ..., f_K^{'})$. Finally, we concatenate $s_i$ with $F_i$ and use a transformation matrix $W_v$ to transform them to the original dimension as $s_i = W^v[s_i, F_i]$. Here, we reuse $s_i$ to represent the refined visual-contextualized semantic node representation for simplicity.

	\noindent
	\textbf{Hierarchical Semantics Aggregation.} Based on the insight that semantic events are composed of a sequence of interactional actions and objects, we propose the hierarchical semantic aggregation mechanism, which aggregates the semantics from the contextualized action nodes and object nodes to compose the global event nodes. Inspired by the success of positional query encoding~\cite{carion2020end} in object detection, we initialize the event nodes as a set of learnable query vectors $\{p_i\}_{i=1}^{N_p}$ and then aggregate relevant semantics from action nodes and object nodes to refine the event nodes. Here we take the video semantic graph as an illustration. For an event query $p_i$, we calculate the attention weights over semantic nodes $\{s_j\}_{j=1}^{N_v}$ and update the $p_i$, given by:
	
	\begin{equation}
	\tilde{p_i}\!=\!\sum_{j=1}^{N_v} \alpha_{ij}^{e}\cdot s_j, \alpha_{ij}^{e}\!=\!\frac{exp({(W^e_1 p_i)}^T \cdot (W^e_2 s_j))}{\sum_{j=1}^{N_v} exp({(W^e_1 p_i)}^T \cdot (W^e_2 s_j))}
	\end{equation}
	
	\noindent
	where $W^e_1, W^e_2$ are projection matrices, and $\tilde{p_i}$ is the semantics-aware event node. Subsequently, we stack multiple such graph self-attention layers and merge the final event nodes into $\{s_j\}_{j=1}^{N_v}$ to form the complete hierarchical semantic graph of video, denoted by $M = \{m_i\}_{i=1}^{N_m} \in \mathbb{R}^{d \times N_m}$. In the same manner, we can obtain the complete hierarchical semantic graph of language, denoted by $H = \{h_i\}_{i=1}^{N_h} \in \mathbb{R}^{d \times N_h}$. $M$ and $H$ are the unified structure of three semantic hierarchies, which tightly couple multi-granularity semantics between the two modalities.

	\subsection{Variational Cross-Graph Correspondence}\label{s4.2}
	After parsing both videos and language queries into individual hierarchical semantic graphs, we then model the cross-modality interactions between two graphs by cross-graph convolution, and induce the fine-grained semantic correspondence between them for final prediction. The objective function can be formulated as $P(Y|M, H)$, where $Y = (t^s, t^e)$ represents the target time interval. Since the ground-truth correspondence between the nodes of two graphs is not available, we treat the cross-graph correspondence as a latent variable $z$. The problem can then be formulated into a variational inference framework~\cite{sohn2015learning} and the objective function can be rewritten as $P(Y|M, H, z)P(z|M, H)$. Instead of directly maximizing $P(Y|M, H)$, we propose to maximize its evidence lower bound~(ELBO)~\cite{kingma2013auto} as follows:
	
	\begin{equation}\label{e5}
	\begin{aligned}
	ELBO (\phi, \theta) \quad = & \quad E_{q_{\phi}(z|M, H, Y)}log p_{\theta}(Y|M, H, z) \\
	&- KL(q_{\phi}(z|M, H, Y) || p_{\theta}(z|M, H)) \\
	&\leq log p(Y|M, H)
	\end{aligned}
	\end{equation}
	\noindent
	Specifically, we characterize $P(Y|M, H)$ using three components: a prior model $p_{\theta}(z|M, H)$, a posterior model $q_{\phi}(z|M, H, Y)$, and a likelihood model $p_{\theta}(Y|M, H, z)$. In the following, we first introduce cross-graph convolution to model the semantic interaction between two graphs and then elaborate these three models in detail.
	
	\noindent
	\textbf{Cross-Graph Convolution.} Given the graphs $M$ and $H$, the cross-graph convolution is performed between the nodes in the same hierarchical levels. For a video semantic node $m_i^k$, the cross-convolution from $H$ to $M$ is formulated as:
	
	\begin{equation}
	\alpha_{ij}^{h2m} = \frac{exp({(W^c_1 m_i^k)}^T \cdot (W^c_2 h_j^k))}{\sum_{j \in \mathcal{N}_H^k} exp({(W^c_1 m_i^k)}^T \cdot (W^c_2 h_j^k))}
	\end{equation}
	
	\begin{equation}
	\tilde{m}_i^k = (1 - \beta_i^k)\odot m_i^k  +  \beta_i^k \odot \sum_{j \in \mathcal{N}_H^k} \alpha_{ij}^{h2m} \cdot h_j^k, k \in \{e, a, o\}
	\end{equation}
	
	\noindent
	where the gate unit $\beta_i^k\!=\!\sigma(U^gm_i^k + b)$ controls the information flow from $H$ to $M$, $k$ denotes three semantic levels (\textsl{i.e.,} \textsl{event, action, object}), $\mathcal{N}_H^k$ denotes the nodes of $H$ in level $k$. In a similar manner but reversed order, we can obtain $\tilde{H}$.
	
	\noindent
	\textbf{Prior Model.} Taking $\tilde{M}$ and $\tilde{H}$ as input, the prior model $p_{\theta}(z|M, H)$ is to infer the cross-graph correspondence captured by a latent variable $z \in \mathbb{R}^{N_m \times N_h}$, where $z_{ij}$ corresponds to the semantic correspondence between $\tilde{m}_i$ and $\tilde{h}_j$. 
	Specifically, the $z_{ij}$ can be formulated as:
	
	\begin{equation}
	\tilde{z}_{ij} = {(W^s_1 \tilde{m}_i)}^T \cdot (W^s_2 q \odot \tilde{h_j}), z_{ij} = \frac{exp(\tilde{z}_{ij})}{\sum_{j=1}^{N_h} exp(\tilde{z}_{ij})}
	\end{equation}
	
	\noindent
	where $q$ is the global sentence feature that guides the semantic correspondence inference.
	
	\noindent
	\textbf{Approximate Posterior Model.} The posterior model $q_{\phi}(z|M, H, Y)$ infers the cross-graph correspondence by additionally considering the information of ground-truth $Y$. According to the temporal boundary $Y$, we can determine the segments in $Y$ and the action and object nodes that correspond to these segments. These nodes in the video graph contain the most relevant semantics to the language semantic graph, which can better guide cross-graph correspondence learning. Therefore, we obtain $m^*$ through \verb|Avg-Pooling| over these nodes and use $m^*$ to guide the correspondence learning:
	
	\begin{equation}
	\tilde{z}_{ij} = {(W^s_3 m^* \odot \tilde{m}_i)}^T \cdot (W^s_4 q \odot \tilde{h_j}),\ z_{ij} = \frac{exp(\tilde{z}_{ij})}{\sum_{j = 1}^{N_h} exp(\tilde{z}_{ij})}
	\end{equation}
	
	\noindent
	where $m^*$ and $q$ serve as global visual and linguistic guidance, respectively.
	
	\begin{figure*}[!t]
		\centering
		\includegraphics[width=0.85\textwidth]{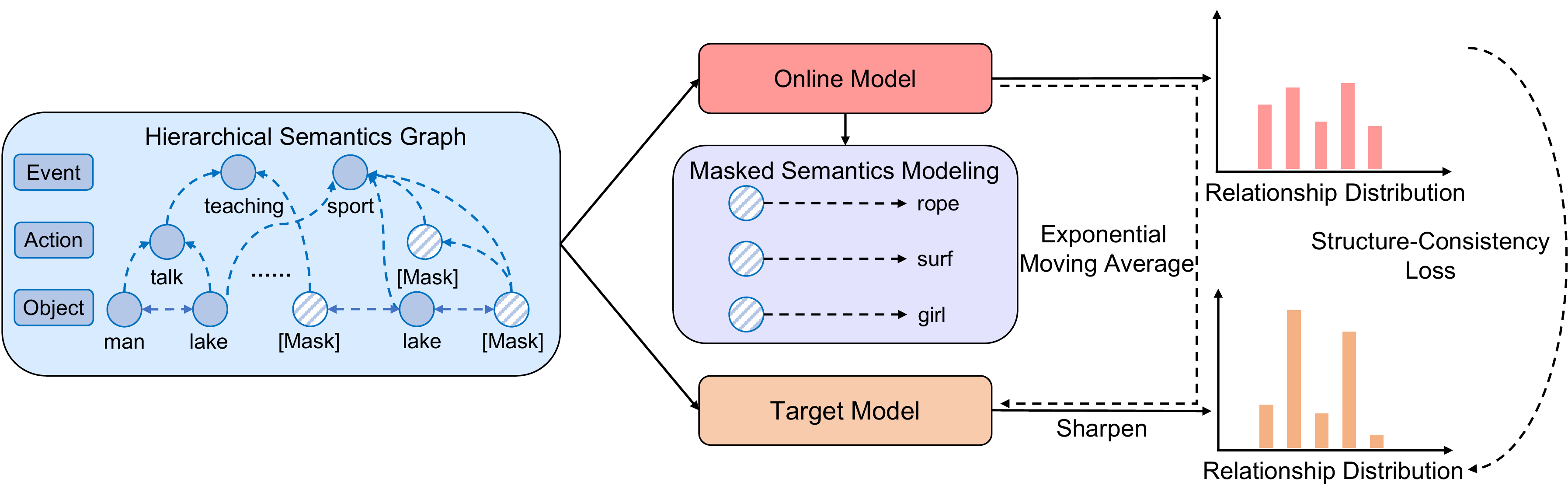}
		\vspace{-0.2cm}
		\caption{Overview of the structure-informed semantics modeling pipeline (Section~\ref{s5.1}).}
		\label{SISM}
		\vspace{-0.2cm}
	\end{figure*}
	
	\noindent
	\textbf{Likelihood Model.} The likelihood model $p_{\theta}(Y|M, H, z)$ predicts the temporal boundary based on the latent correspondence $z$ and hierarchical semantic graphs $\tilde{M}$ and $\tilde{H}$. Concretely, to obtain joint multi-modality representations, we first integrate two graphs guided by the learned cross-graph correspondence:
	
	\begin{equation}
	M^{'} = z \tilde{H} \in \mathbb{R}^{d \times N_m}, \quad M^J = W^J [\tilde{M}; M^{'}] \in \mathbb{R}^{d \times N_m}
	\end{equation}
	
	\noindent
	where projection matrix $W^J\!\in\!\mathbb{R}^{d \times 2d}$ and $M^J$ is the joint multi-modality representations of the hierarchical semantic graph. Next, we use $M^J$ to refine segment features $X = \{x_t\}_{t=1}^T \in \mathbb{R}^{d \times T}$. We adopt \verb|Avg-Pooling| over frame features $\{f_i^t\}_{i=1}^K$ of segment $V_t$ to compute the segment features $x_t$. Subsequently, we adopt multi-head cross-modal attention to softly select relevant information from $M^J$ to $X$. Concretely, we take $X$ as queries and $M^J$ as keys and values:
	
	\begin{equation}
	X^* = MultiAttn(X, M^J, M^J)
	\end{equation}
	
	\noindent
	where $X^*$ is the semantics-aware segment representations. Then, we summarize the segment representations using \verb|Attentive-Pooling| based on the sentence feature $q$:
	
	\vspace{-0.5cm}
	\begin{equation}
	v^* = \sum_{i=1}^T \alpha^q_{i} \cdot x^*_i, \quad\alpha^q_{i} = \frac{exp((W^q_1q)^T\cdot(W^q_2x_i^*))}{\sum_{i=1}^T exp((W^q_1q)^T\cdot(W^q_2x_i))}
	\end{equation}
	
	\noindent
	where $v^*$ is the summarized video feature. Finally, we predict the time interval $Y = (t^s, t^e)$ as $t^s, t^e = MLP(v^*)$.
	
	\subsection{Optimization}
	We formulate the objective function as:
	
	\begin{equation}
		\mathcal{L}_{ELBO}(\phi, \theta) = - ELBO(\phi, \theta)
	\end{equation}
	Through minimizing the $\mathcal{L}_{ELBO}$, we maximize the evidence lower bound of $P(Y|M, H)$. According to Equation \ref{e5}, $\mathcal{L}_{ELBO}$ consists of two terms. The first term corresponds to the regression loss. Specifically, following \cite{mun2020local}, we minimize the sum of smooth $L_1$ distances between the normalized ground-truth time interval $(\hat{t^s}, \hat{t^e}) \in [0, 1]$ and our prediction $(t^s, t^e)$. This term not only teaches the likelihood model to predict the correct time interval but also encourages the approximate posterior model to learn more accurate cross-graph correspondence. The second term corresponds to the KL-divergence loss. Concretely, as the latent variable $z$ is a correlation matrix, we compute the KL-divergence by rows. Intuitively, through minimizing this term, we can teach the prior model to capture the cross-graph semantic correspondence as well as the approximate posterior model. During inference without access to the ground-truth, we can replace the approximate posterior model with the learned prior model to infer the cross-graph correspondence. Notably, the posterior model is employed to generate $z$ during training.

	\section{Adaptive Structured Semantics Learning}\label{s5}
\begin{figure*}[!t]
		\centering
		\includegraphics[width=0.85\textwidth]{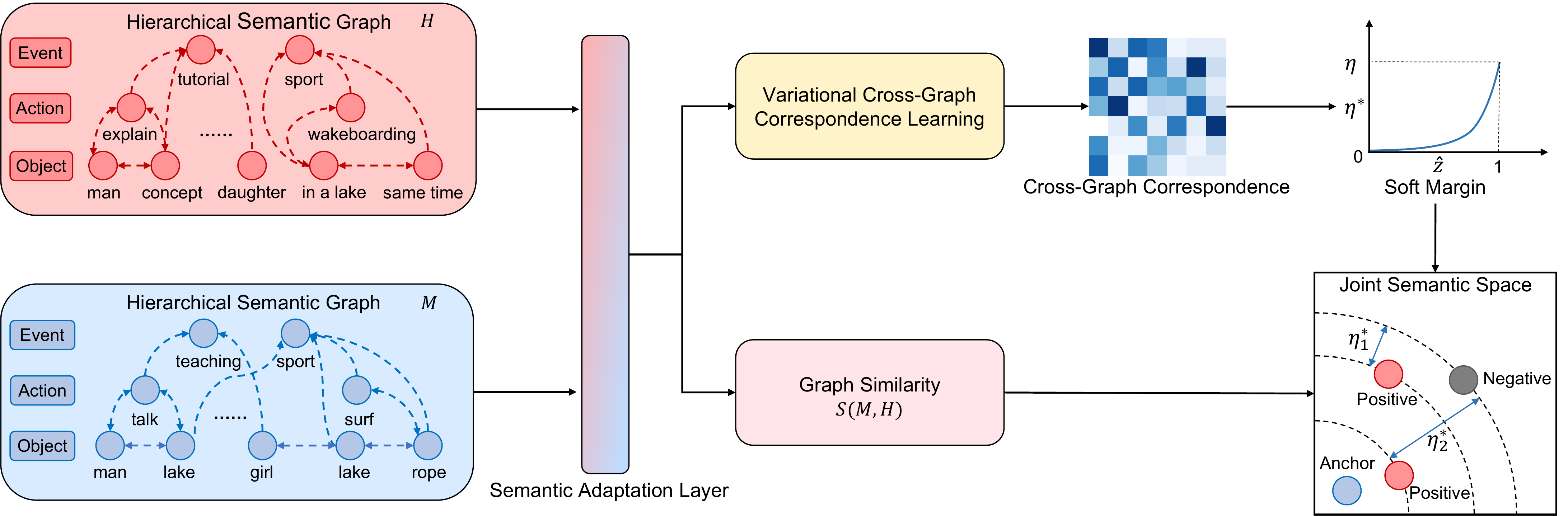}
		\vspace{-0.2cm}
		\caption{Overview of the cross-graph semantic adaptation pipeline (Section~\ref{s5.2}).}
		\label{CGSA}
		\vspace{-0.2cm}
	\end{figure*}

	In this section, we newly propose an adaptive structured semantics learning (ASSL) approach to encourage the VISA model to learn structure-informed and domain-generalizable semantics while simultaneously addressing the semantic node noise and cross-graph semantic shift problems of the VISA. As shown in Figure~\ref{SISM} and Figure~\ref{CGSA}, ASSL includes two learning objectives: \textbf{structure-informed semantics modeling}~(Section~\ref{s5.1}) and \textbf{cross-graph semantic adaptation}~(Section~\ref{s5.2}).

	\subsection{Structure-Informed Semantics Modeling}\label{s5.1}
	We first present the intuitive version of masked semantics modeling for the video hierarchical semantic graph and then introduce a structure-informed self-training strategy to effectively investigate the structure information of the video hierarchical semantic graph in a robust manner.

	\noindent
	\textbf{Masked Semantics Modeling.} To motivate the proposed hierarchical semantic graph to capture relevant visual context from videos and diverse semantic context from the interaction between semantic nodes, we design the masked semantics modeling as an additional objective. Specifically, we randomly mask out the action nodes and object nodes in the video hierarchical semantic graph with a probability of 10\%, and replace the masked node $\overline{s}_i^{\bullet} \in \{\overline{S}^a, \overline{S}^o\}$ with a special node \verb|[MASK]|. After semantic- and visual- contextualized learning, we obtain the contextualized representations of the masked node $s_i^{\bullet}$, which aggregates the semantic context from neighboring semantic nodes and visual context from the video. Then, the goal is to predict the corresponding semantic label~(\textsl{i.e., action or object classes predicted by the off-the-shelf detection models}) based on the contextualized masked node representation $s_i^{\bullet}$, by minimizing the negative log-likelihood:
	
	\begin{equation}
	\mathcal{L}_{MSM} = -\frac{1}{|S^{\bullet}|} \sum_{s_i^{\bullet} \in S^{\bullet}} log P(Y^{MSM}_i|s_i^{\bullet})
	\end{equation}
	where $S^{\bullet}$ is the set of the masked semantic nodes, and $Y^{MSM}_i$ is the one-hot vector that corresponds to the action or object classes extracted by the detection models.
	
	\noindent
	\textbf{Structure-Informed Self-Training.} Considering the inevitable noise from the detection models, $Y^{MSM}_i$ might not always be reliable and could degrade the generalizability of the model. Instead of directly fitting the model to the one-hot labels, we propose a structure-informed self-training strategy based on a moving-averaged target network, which generates structure-informative pseudo targets to prevent the model from being over penalized by the noisy labels. Specifically, the target network has the same architecture as the current online VISA model but uses a different set of parameters $\xi$. Let $\theta$ denote the parameters of the current online VISA model, and then the target network is updated by exponential moving average (EMA)~\cite{grill2020bootstrap}:
	\begin{equation}
	\xi \leftarrow \omega \xi + (1 - \omega) \theta
	\end{equation}
	where $\omega \in [0, 1]$ is a target decay rate.
	
	For each masked node representation $s_i^{\bullet}$ from the target network, we compute its similarity with other nodes to obtain the relationship distribution $P_i^T \in \mathbb{R}^{N_m}$, where $P_{ij}^T$ is computed as:
	
	\begin{equation}
	P_{ij}^T = \frac{exp(s_i^{\bullet} \cdot s_j)}{\sum_{j=1}^{N_m} exp(s_i^{\bullet} \cdot s_j)}
	\end{equation}
	
	In the same manner, we can obtain the relationship distribution $P_i^O$ from the online VISA model. Then, we take $P_i^T$ as the soft pseudo targets and propose to push the structure consistency between $P^T$ and $P^O$ by minimizing the Kullback-Leibler divergence. The structure-consistency loss is then defined as:
	
	\begin{equation}
	\mathcal{L}_{SC} = D_{KL}(P^T||P^O) \label{e_lsc}
	\end{equation}

	Compared with the one-hot labels, the proposed pseudo targets are more informative as it contains more diverse semantic structure information between the masked nodes and other nodes in the graph. Taking pseudo targets as additional supervision, the VISA model is incentivized to better encode the structured semantics and learn structure-informed representations.
	
	\noindent
	\textbf{Sharpening Relationship Distribution.} To encourage the model to focus on the most significant semantic relationships, we design a sharpening function over the target relationship distribution, which can reduce the entropy of the target relationship distribution. The sharpened target relationship distribution is defined as:
	
	\begin{equation}
	\tilde{P}^T_{ij} = \frac{{P^T_{ij}}^{\frac{1}{\tau}}}{\sum_{j=1}^{N_m} {P^T_{ij}}^{\frac{1}{\tau}} }
	\end{equation}
	where $\tau$ is a temperature hyperparameter. By lowering the temperature, the target distribution will be sharper.
	
	By replacing $P^T$ in Equation \ref{e_lsc} with the sharpened $\tilde{P}^T$, the final objective for structure-informed semantics modeling can be formulated as the weighted combination of the structure-consistency loss and the intuitive masked semantics modeling loss:
	
	\begin{equation}
	L_{SISM} = (1 - \lambda) \mathcal{L}_{MSM} + \lambda \mathcal{L}_{SC}
	\end{equation}

	\subsection{Cross-Graph Semantic Adaptation}\label{s5.2}
	As the semantic nodes of video and language graphs are initialized from different sources (\textsl{i.e., action and object class labels from detection models, verb and noun phrases from query sentences}), there is a semantic shift problem between the two graphs. For instance, the object node ``\textsl{running}'' in the video graph might be represented as \textsl{``rush'' or ``dash''} in the language graph. Such semantic shift between video and language might hinder the cross-graph correspondence reasoning. To address the cross-graph semantic shift problem, we propose a cross-graph semantic adaptation objective, where a semantic adaptation layer supervised by a semantic-adaptive margin loss is designed to derive domain generalizable representations.

	\noindent
	\textbf{Semantic Adaptation Layer.} We insert a linear semantic adaptation layer between the hierarchical semantic graph and variational cross-graph correspondence reasoning. Formally, given the learned hierarchical semantic graphs $M \in \mathbb{R}^{d \times N_m}$ and $H \in \mathbb{R}^{d \times N_h}$, we adopt linear transformation matrices $W^A_1 \in \mathbb{R}^{d \times d}$ and $W^A_2 \in \mathbb{R}^{d \times d}$ on them, respectively: $M = W^A_1M, H = W^A_2H$. Here we reuse $M$ and $H$ for simplicity. 
	
	\begin{table*}[!htb]
		\caption{Performances (\%) of SOTA temporal grounding models and our approach on the proposed Charades-CG datasets.}
		\label{t2}
		\resizebox{\textwidth}{!}{
			\centering
			\begin{tabular}{ ll cccc cccc cccc}
				\hline
				\multicolumn{2}{l}{\multirow{2}*{Method}} 
				&\multicolumn{4}{c}{\textsl{Test-Trivial}} 
				&\multicolumn{4}{c}{\textsl{Novel-Composition}}
				&\multicolumn{4}{c}{\textsl{Novel-Word}}
				\\
				
				\cmidrule(lr){3-6} \cmidrule(lr){7-10}  \cmidrule(lr){11-14} 
				\multicolumn{2}{l}{} &IoU=0.3 &IoU=0.5 &IoU=0.7 &\multicolumn{1}{c}{mIoU}  &IoU=0.3 &IoU=0.5 &IoU=0.7 &\multicolumn{1}{c}{mIoU} &IoU=0.3 &IoU=0.5 &IoU=0.7 &\multicolumn{1}{c}{mIoU}  \\

				\multicolumn{2}{l}{WSSL}  
				&27.25  &15.33  &5.46    &18.31   &14.19   &3.61 &1.21 &8.26  &15.43  &2.79  &0.73  &7.92  \\

				\multicolumn{2}{l}{TSP-PRL}  
				&57.69  &39.86 & 21.07   &38.41    &32.18  &16.30  &2.04 &13.52   &36.74  &14.83  &2.61  &14.03   \\

				\multicolumn{2}{l}{TMN} 
				&34.19  &18.75    &8.16  &19.82  &21.37  &8.68    &4.07  &10.14  &23.01  &9.43    &4.96  &11.23 \\

				\multicolumn{2}{l}{2D-TAN} 
				&63.99  &48.58 &26.49  &44.27 &46.60  &30.91  &12.23  &29.75 &42.05  &29.36  &13.21  &28.47 \\

				\multicolumn{2}{l}{LGI} 
				&66.89  &49.45  &23.80    &45.01  &44.22  &29.42 &12.73   &30.09  &40.43  &26.48  &12.47 &27.62\\

				\multicolumn{2}{l}{VLSNet}  
				&61.60  &45.91  &19.80   &41.63   &38.29  &24.25  &11.54 &31.43 &40.07  &25.60  &10.07 &30.21  \\
				
				\hline
				\multicolumn{2}{l}{\textbf{Ours}}  
				&   &    &      &    &   &   &    &  &   &  &   &  \\
				
				\multicolumn{2}{l}{VISA}  
				&72.84  &53.20  &26.52    &47.11  &60.81  &45.41 &22.71 &42.03 & 61.28 &42.35 &20.88 &40.18  \\
				
				\multicolumn{2}{l}{VISA + ASSL}  
				&\textbf{76.31}   &\textbf{56.14}     &\textbf{28.27}       &\textbf{48.92}     &\textbf{64.57}    &\textbf{47.76}    &\textbf{24.85}     &\textbf{43.89}   &\textbf{65.32}    &\textbf{44.75}   & \textbf{22.31}   &\textbf{42.38}   \\
				
				\hline
			\end{tabular}
		}
	\end{table*}

	\begin{table*}[!htb]
		\caption{Performances (\%) of SOTA temporal grounding models and our approach on the proposed ActivityNet-CG datasets.}
		\label{t3}
		\resizebox{\textwidth}{!}{
			\centering
			\begin{tabular}{ ll cccc cccc cccc}
				\hline
				\multicolumn{2}{l}{\multirow{2}*{Method}} 
				&\multicolumn{4}{c}{\textsl{Test-Trivial}} 
				&\multicolumn{4}{c}{\textsl{Novel-Composition}}
				&\multicolumn{4}{c}{\textsl{Novel-Word}}
				\\
				
				\cmidrule(lr){3-6} \cmidrule(lr){7-10}  \cmidrule(lr){11-14} 
				\multicolumn{2}{l}{} &IoU=0.3 &IoU=0.5 &IoU=0.7 &\multicolumn{1}{c}{mIoU}  &IoU=0.3 &IoU=0.5 &IoU=0.7 &\multicolumn{1}{c}{mIoU} &IoU=0.3 &IoU=0.5 &IoU=0.7 &\multicolumn{1}{c}{mIoU}  \\
				\hline

				\multicolumn{2}{l}{WSSL}  
				&24.68  &11.03 &4.14    &15.07 &11.07 &2.89  &0.76 &7.65 &12.60 &3.09 &1.13 &7.10 \\

				\multicolumn{2}{l}{TSP-PRL}  
				&52.38 &34.27   &18.80    &37.05 &29.75  &14.74 &1.43  &12.61 &34.11  &18.05 &3.15 &14.34 \\

				\multicolumn{2}{l}{TMN} 
				&32.20   &16.82   &7.01 &17.13 &22.56   &8.74   &4.39 &10.08  &23.76   &9.93   & 5.12 &11.38 \\

				\multicolumn{2}{l}{2D-TAN} 
				&57.45   &44.50	&26.03	&42.12 &33.06   &22.80  &9.95  &28.49 &33.76    &23.86 &10.37 &28.88 \\

				\multicolumn{2}{l}{LGI} 
				&60.13   &43.56 &23.29  &41.37 &34.10   &23.21  &9.02  &27.86 &32.43   &23.10 &9.03 &26.95\\

				\multicolumn{2}{l}{VLSNet}  
				&57.15   &39.27   &23.12     &42.51  &31.47   &20.21 &9.18 &29.07 &31.95   &21.68 &9.94 &29.58\\
				
				\hline
				\multicolumn{2}{l}{\textbf{Ours}}  
				&   &    &      &    &   &   &    &  &   &  &   &  \\
				
				\multicolumn{2}{l}{VISA}  
				&64.03  &47.13  &29.64    &44.02  &47.10  &31.51 &16.73 &35.85 &46.24 &30.14 &15.90 &35.13  \\
				
				\multicolumn{2}{l}{VISA + ASSL}  
				&\textbf{67.46}   &\textbf{49.37}     &\textbf{31.18}       &\textbf{46.15}     &\textbf{49.62}    &\textbf{33.22}    &\textbf{17.83}     &\textbf{37.56}   &\textbf{48.73}    &\textbf{32.04}   & \textbf{17.24}   &\textbf{36.87}   \\
				\hline
				
			\end{tabular}
			
		}
	\end{table*}

	\noindent
	\textbf{Semantic-Adaptive Margin Loss.} As the query sentence refers to a target segment of the video rather than the whole video, the corresponding language graph only matches a part of the video graph. Therefore, it might not be suitable that directly minimize the discrepancy of representations between two graphs following existing domain adaptation methods~(\textsl{e.g.,} maximum mean discrepancy~\cite{gretton2012kernel}, domain adversarial neural networks~\cite{ajakan2014domain}, contrastive learning~\cite{oord2018representation}). Alternatively, we consider a ranking-based loss formalism with a soft margin, which is adaptively determined according to the semantic overlap degree between the two graphs. Specifically, the semantic-adaptive margin loss is formulated as:

	\begin{equation}
	\begin{aligned}
	\mathcal{L}_{SAM} = &\quad [\eta^* - \mathcal{S}(M,H) + \mathcal{S}(M, \overline{H})]_{+} \\
	&+ [\eta^* - \mathcal{S}(M,H) + \mathcal{S}(\overline{M},H)]_{+}
	\end{aligned}
	\end{equation}
	where $\eta^*$ is the soft margin, $(M, H)$ is the paired input video and query, $\overline{M}$ and $\overline{H}$ is the mismatched negative samples from the current batch, and $[x]_+ = max(x, 0)$. The $\mathcal{S}(M, H)$ represents the similarity score between two graphs. For $\mathcal{S}(M, H)$, we first compute the global graph representation for each graph by performing \verb|Avg-Pooling| over their nodes. Then, we compute the similarity between the global representations of two graphs. 
	
	Intuitively, the more semantics shared between two graphs, the larger the margin should be. The soft margin $\eta^*$ is then adaptively determined by:
	\begin{equation}
	\eta^* = \frac{u^{\hat{z}} - 1}{u - 1}\eta
	\end{equation}
	where $u$ is the curve parameter, $\hat{z} \in (0, 1)$ reflects the semantic overlap degree between two graphs, and $\eta$ is the base margin hyper-parameter. if $\hat{z}$ is close to 1, $\eta^*$ will be assigned a larger value, and a small value otherwise. 
	
	Given the inferred cross-graph correspondence $z \in \mathbb{R}^{N_m \times N_h}$, where $z_{ij}$ represents the alignment score between the \textsl{i-th} video graph node and the \textsl{j-th} language graph node, we first calculate the maximum alignment score in row-wise for each video graph node, and then the semantic overlap degree $\hat{z}$ is obtained by averaging the maximum alignment score over all video graph nodes:
	
	\begin{equation}
	\hat{z} = \verb|Avg-Pooling|(\hat{z_1}, ..., \hat{z_i}, ..., \hat{z_{N_m}}),\ \hat{z_i} = max_j z_{ij}
	\end{equation}
	If the video shares a larger part of semantics with the query, the semantic overlap degree $\hat{z}$ will have a larger value. In this way, we adaptively determine the soft margin to adjust the representations of video graph and language graph into a unified semantic space.
	
\subsection{Optimization}
Combined with the $\mathcal{L}_{ELBO}$ introduced in Section~\ref{s4}, the full objective of  our proposed framework can be formulated as:

\begin{equation}
	\mathcal{L} = \mathcal{L}_{ELBO} + \mathcal{L}_{SISM} + \mathcal{L}_{SAM}
\end{equation}

	\section{Experiments}
	
	\subsection{Datasets and Evaluation Metrics}
	\noindent
	\textbf{Datasets.} We evaluate the compositional generalizability of our method on the proposed Charades-CG and ActivityNetCG datasets, each of which consists of three testing splits: \textsl{Test-Trivial, Novel-composition,} and \textsl{Novel-Word}. The \textsl{Test-Trivial} split is similar to the testing split of original datasets, where the compositions have occurred in the training split. For the \textsl{Novel-Composition} split, each query sentence contains one type of novel composition, which has never been seen in the training split. For the \textsl{Test-Trivial} split, each query sentence contains a novel word.
	
	\noindent
	\textbf{Evaluation Metric.} Following previous works~\cite{zhang2020learning, mun2020local}, we adopt ``R@n, IoU=m''  and mIoU~(\textsl{i.e.,} the average temporal IoU) as our evaluation metrics. Specifically, given a testing query, it first calculates the temporal Intersection-over-Union (IoU) between the predicted moment and the ground truth, and ``R@n, IoU=m'' is defined as the percentage of at least one of top-n predictions with IoU larger than m. We report the top-1 performance with IoU thresholds of $\{0.3, 0.5, 0.7\}$.

	\subsection{Implementation Details}
	For all methods, we use the public official implementations to get their compositional temporal grounding results. We train them on the training split and evaluate them on the test-trivial, novel-composition, and novel-word splits, respectively. Following \cite{yuan2021closer}, we use unified video and language features for more fair comparisons. Concretely, we use I3D features~\cite{carreira2017quo} for the video in Charades-CG and C3D features~\cite{tran2015learning} for the videos in ActivityNet-CG. We use pre-trained GloVe~\cite{pennington2014glove} word vectors to initialize each word in the language queries. 
	
	In our proposed framework, we use the I3D model~\cite{carreira2017quo} pretrained on kinetics~\cite{kay2017kinetics} dataset as our action detector, and use Faster R-CNN with ResNet-101~\cite{ren2015faster, Anderson2017up-down,he2016deep} pre-trained on  Visual Genome~\cite{krishnavisualgenome} dataset as our object detector. For an untrimmed video, we divide it into a sequence of segments with a fixed length~(\textsl{i.e.,} 32 frames), and then adopt the off-the-shelf object and action detectors to extract objects and actions for each segment. For each segment, we select the top-3 action classes and top-5 object classes with the highest confidence score as action nodes and object nodes, respectively. The dimension of input video features is 1024 and the dimension of GloVe~\cite{pennington2014glove} vectors is 300. We set the dimension of all node~(three hierarchies of the two graphs) representations as 384. The target decay rate $\omega$ for the moving-averaged target network is set to 0.995. The temperature hyper-parameter $\tau$ for relationship distribution sharpening is set to 0.5. We set the base margin hyper-parameter $\eta$ to 0.4 and the curve parameter $u$ to 10. During training, we set the batch size to 32 and use Adam as optimizer~\cite{duchi2011adaptive} with the learning rate of $1e^{-4}$.

	\subsection{Comparison with State-of-the-Art Methods}
	We evaluate the proposed \textbf{VISA + ASSL} approach and compare it with recently proposed state-of-the-art methods, including: 1)~Proposal-based methods: \textbf{TMN}~\cite{liu2018temporal}, \textbf{2D-TAN}~\cite{zhang2020learning}; 2)~Proposal-free methods: \textbf{LGI}~\cite{mun2020local}, \textbf{VLSNet}~\cite{zhang2020span}; 3)~RL-based method: \textbf{TSP-PRL}~\cite{wu2020tree}; 4)~Weakly-supervised method: \textbf{WSSL}~\cite{duan2018weakly}. 
	
	Table \ref{t2} and Table \ref{t3} summarize the results of the above methods on compositional temporal grounding. Overall, our VISA + ASSL achieves the highest performance on all dataset splits, demonstrating the superiority of our proposed model. Notably, we observe that the performance of all tested SOTA models drops significantly on the novel-composition and novel-word splits. The difference in performance between test-trivial and novel-composition~(novel-word) ranges up to 20\%. In contrast, our VISA surpasses them by a large margin on novel-composition and novel-word splits, demonstrating superior compositional generalizability. Particularly, for the novel-composition splits of Charades-CG and ActivityNet-CG datasets, our VISA+ASSL significantly surpasses the SOTA methods by 30.86\% and 23.32\%  relatively on mIoU, respectively.
	
	Moreover, the newly proposed adaptive structured semantics learning (ASSL) approach further improves the VISA model by enforcing the VISA model to capture structure-informed and domain-generalizable semantics from video and language. Comparing VISA + ASSL with VISA, our ASSL consistently improves the performance over all metrics. Particularly, for the novel-composition splits of Charades-CG and ActivityNet-CG datasets, the ASSL improves the performance of VISA with 1.86\% and 1.71\% absolute performance gains on mIoU, respectively.
	
	The main reasons for our proposed approach outperforming the previous competing models lie in three folds. First, the hierarchical semantic graph explicitly models the structured semantics of video and language, bridging the semantic gap between the two modalities. Second, the variational cross-graph correspondence reasoning establishes fine-grained semantic correspondence between two modalities, allowing for joint compositional reasoning. Third, the adaptive structured semantics learning promotes the learned representations of two graphs to be more structure-informed and domain-generalizable, thus facilitating the cross-graph correspondence reasoning.
	

	\begin{figure*}[!t]
		\centering
		\includegraphics[width=\textwidth]{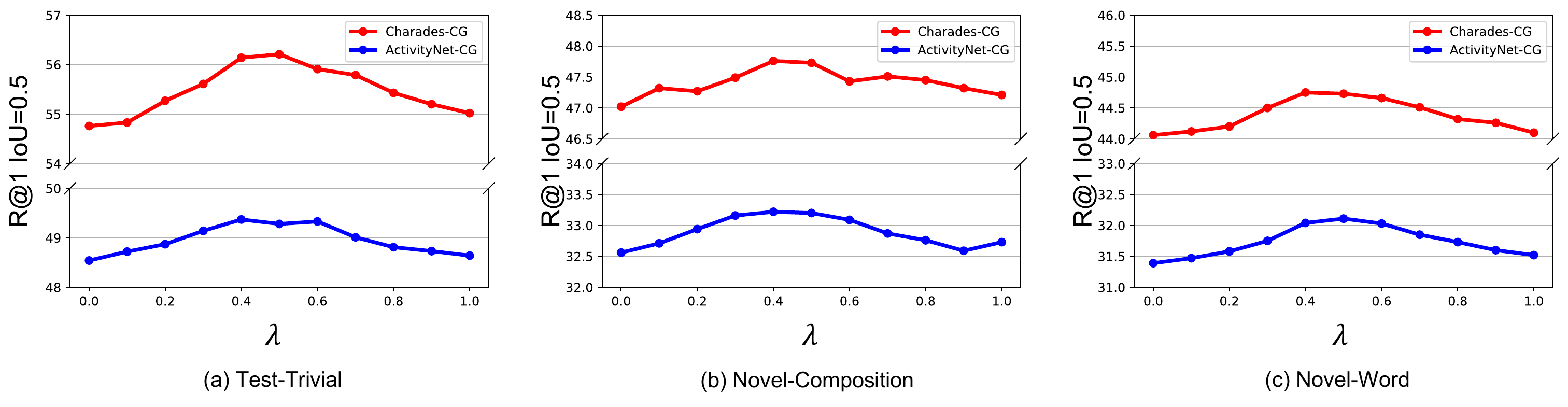}
		\caption{Performance (R@1, IoU=0.5) comparison with respect to different $\lambda$ values.}
		\label{ablation_lambda}
	\end{figure*}

	\begin{table}[!t]
		\caption{Ablation study of VISA with metric R@1, IoU=0.5 on test-trivial~(Trivial),  novel-composition~(Comp) and novel-word~(Word) splits.}
		\label{t4}
		\centering
		\begin{tabular}{ lll|ccc|ccc}
			\hline
			&\multicolumn{2}{l|}{\multirow{2}*{Method}} 
			&\multicolumn{3}{c|}{Charades-CG} 
			&\multicolumn{3}{c}{ActivityNet-CG}
			\\
			
			&\multicolumn{2}{l|}{} &Trivial &Comp &\multicolumn{1}{c|}{Word}  &Trivial &Comp &Word  \\
			\hline    
			1 &\multicolumn{2}{l|}{w/o SCL}   &50.71 &43.75 &40.16    &44.09   &29.03 &29.41                   \\
			2 &\multicolumn{2}{l|}{w/o VCL}   &49.62 &42.26 &38.62    &44.71  &29.34 &28.09					\\
			3 &\multicolumn{2}{l|}{w/o HSA}  &51.47 &44.22 &41.09   &45.28  &30.29	&29.31	\\
			4 &\multicolumn{2}{l|}{w/o VCC}  &48.82 &41.08  &37.54  &42.64 &27.32 &26.37 \\
			5 &\multicolumn{2}{l|}{Detection} &34.50 &12.97  &11.70 &30.95 &10.92 &10.07					\\
			\hline
			6 &\multicolumn{2}{l|}{{\small \textbf{VISA}}} &\textbf{53.20} &\textbf{45.41} &\textbf{42.35} &\textbf{47.13} &\textbf{31.51} &\textbf{30.14}					\\
			\hline
		\end{tabular}
	\end{table}

	\subsection{Ablation Study}
	
	\subsubsection{Effectiveness of VISA}
	In this section, we conduct an ablation study to illustrate the effectiveness of our variational cross-graph reasoning model in Table~\ref{t4}. Specifically, we train the following ablation models. 1) w/o SCL: we remove the Semantic-Contextualized Learning~(SCL). 2) w/o VCL: we remove the Visual-Contextualized Learning~(VCL). 3) w/o HSA: we remove the Hierarchical Semantics Aggregation~(HSA). 4) w/o VCC: we replace the Variational Cross-graph Correspondence learning~(VCC) by directly using cross-modal self-attention to fuse two graphs. 5) Detection-based: we directly use the detection results and SRL labels as features.
	
	The results of Row 1 and Row 2 indicate that learning fine-grained contextualized information is crucial for compositional reasoning. Also, the results of Row 3 validate the importance of event-level hierarchy on global semantic understanding. Ours w/o VCC does not achieve satisfying results, because directly fusing the graphs of video and sentence could possibly disrupt the semantic correspondence between them, which causes a negative effect on temporal grounding performance. In contrast, the proposed VCC establishes fine-grained cross-graph correspondence by variational inference, which is meticulous and achieves the best results. Furthermore, Row 5 suggests that the main performance gain does not directly come from the pre-trained detection models. Instead, these detected semantic labels serve as unified symbols for joint compositional reasoning.

	\subsubsection{Effectiveness of ASSL}
	In this section, we verify the effectiveness of our adaptive structured semantics learning, which includes the basic masked semantics modeling loss $\mathcal{L}_{MSM}$, the structure-consistency loss $\mathcal{L}_{SC}$, and the semantic-adaptive margin loss $\mathcal{L}_{SAM}$. We take the VISA model as the backbone~(Row 1) and gradually add them to generate ablation models. As shown in Table~\ref{t5}, the full model (VISA + ASSL, Row 8) outperforms all ablation models, validating each loss is helpful for compositional temporal grounding. In more detail, we observe the following. First, masked semantics modeling loss $\mathcal{L}_{MSM}$ enhances the hierarchical semantic graph to capture more semantic and visual context information, improving the performance from 45.40\% to 46.37\% (novel-composition splits of Charades-CG), and from 31.51\% to 32.09\% (novel-composition splits of ActivityNet-CG). Second, structure-consistency loss $\mathcal{L}_{SC}$ contributes to 1.01\% and 0.62\% improvement on novel-composition splits of Charades-CG and ActivityNet-CG, respectively, by exploring the structured semantics of hierarchical semantic graphs in a robust self-training manner. Third, semantic-adaptive margin loss $\mathcal{L}_{SAM}$ takes up 1.30\% and 0.75\% gain on novel-composition splits of Charades-CG and ActivityNet-CG, respectively, by addressing the semantic shift between video and language graphs. In the end, the results from Row 5 to Row 7 suggest that the proposed losses can enhance the VISA model by deriving structure-informed and domain-generalizable graph representations in a mutually rewarding way.

	\begin{table}[!t]
		\caption{Ablation study of ASSL with metric R@1, IoU=0.5 on test-trivial~(Trivial), novel-composition~(Comp) and novel-word~(Word) splits.}
		\label{t5}
		\setlength\tabcolsep{3pt}
		\centering
		\begin{tabular}{ c|ccc|ccc|ccc}
			\hline
			
			&\multicolumn{1}{l}{\multirow{2}*{$\mathcal{L}_{MSM}$}} 
			&\multicolumn{1}{c}{\multirow{2}*{$\mathcal{L}_{SC}$}}
			&\multicolumn{1}{c|}{\multirow{2}*{$\mathcal{L}_{SAM}$}}
			&\multicolumn{3}{c|}{Charades-CG} 
			&\multicolumn{3}{c}{ActivityNet-CG}
			\\
			
			&\multicolumn{3}{c|}{}  &Trivial &Comp &\multicolumn{1}{c|}{Word}  &Trivial &Comp &Word  \\
			\hline    
			1    &   & &                                            &53.20 &45.40 &42.35  &47.13  &31.51 &30.14   \\
			2&\Checkmark  & &                              &54.12 &46.37 &43.05 &48.08   &32.09 &30.83              \\
			3&&\Checkmark &                                &54.00 &46.41 &43.26 &48.17   &32.13 &30.72              \\
			4&& &\Checkmark                                &54.32 &46.70 &43.47 &48.22   &32.26 &31.09              \\
			
			5&\Checkmark&\Checkmark &             &55.32 &46.86  &43.95 &48.42  &32.68  &31.47              \\
			6&\Checkmark& &\Checkmark             &54.76 &47.02  &44.06 &48.54   &32.56 &31.39              \\
			7&&\Checkmark &\Checkmark             &55.02 &47.21  &44.10 &48.64   &32.73 &31.52              \\
			
			\hline
			8&\Checkmark &\Checkmark  &\Checkmark &\textbf{56.14} &\textbf{47.76}  &\textbf{44.75} &\textbf{49.37} &\textbf{33.22} &\textbf{32.04}		\\
			\hline
		\end{tabular}
	\end{table}

\begin{table}[!t]
	\caption{Performance comparison with respect to different training methods for cross-graph semantic adaptation, with metric R@1, IoU=0.5 on test-trivial~(Trivial), novel-composition~(Comp) and novel-word~(Word) splits.}
	\label{t7}
	\resizebox{\linewidth}{!}{
		\centering
		\begin{tabular}{ ll|ccc|ccc}
			\hline
			\multicolumn{2}{l|}{\multirow{2}*{Method}} 
			&\multicolumn{3}{c|}{Charades-CG} 
			&\multicolumn{3}{c}{ActivityNet-CG}
			\\
			
			\multicolumn{2}{l|}{} &Trivial &Comp &\multicolumn{1}{c|}{Word}  &Trivial &Comp &Word  \\
			\hline    
			
			\multicolumn{2}{l|}{MMD}   &55.67 &46.71    &44.02   &47.82  &32.41  & 31.17                  \\
			\multicolumn{2}{l|}{CL}       &55.97 &47.21    &44.10   &48.63   &32.84 &31.50\\
			\multicolumn{2}{l|}{GAN}    &54.14 &46.28    &43.41   &47.31   &32.05 &30.48				\\
			\multicolumn{2}{l|}{\textbf{Ours - SAM}}    &\textbf{56.14} &\textbf{47.76}  &\textbf{44.75} &\textbf{49.37} &\textbf{33.22} &\textbf{32.04}	 \\
			
			\hline
		\end{tabular}
	}
\end{table}

	\begin{table*}[!htb]
	\caption{Performance comparison on each composition type with metric R@1, IoU=0.5.}
	\label{t6}
	\setlength\tabcolsep{4pt}
	\centering
	\begin{tabular}{ l| ccccc| ccccc}
		\hline
		\multicolumn{1}{l|}{\multirow{2}*{Method}} 
		&\multicolumn{5}{c|}{Charades-CG} 
		&\multicolumn{5}{c}{ActivityNet-CG}\\

		\multicolumn{1}{l|}{}  &Verb-Noun  &Adj-Noun  &Noun-Noun &Verb-Adv   &Prep-Noun &Verb-Noun  &Adj-Noun  &Noun-Noun &Verb-Adv   &Prep-Noun\\				
		\hline    
		2D-TAN     &29.74 &30.09 &31.28 &32.45 &30.17 &22.13  &23.47  &23.64  & 23.68  &23.71 \\
		LGI              &29.68 &29.03 &30.09 &29.31 &29.03 &22.69 &23.70 &23.73   & 23.76  &23.81 \\
		\hline
		VISA   &41.37 &45.06 &43.41 &47.83 &48.61 &28.89   &30.67  &33.93  &35.60  &37.35   \\
		VISA + ASSL &44.84 &46.92 &44.17 &49.28 &51.03 &30.25 &32.48 &36.12 &37.34 &40.08 \\
		
		\hline
	\end{tabular}
\end{table*}
	
	\subsubsection{Impact of Structure-Informed Self-Training}
	In this section, we investigate the impact of the structure-informed self-training by varying the weight hyper-parameter $\lambda$ between masked semantics modeling loss and structure-consistency loss. The larger $\lambda$ indicates that we take more consideration on the soft pseudo targets generated by the moving-averaged target network. As shown in Figure~\ref{ablation_lambda}, when the $\lambda$ is increased from 0 to 0.4, our structure-informed self-training can improve the performance by exploring more structure informative targets and preventing the model from being over penalizing by the noise of masked semantics modeling. We observe that setting $\lambda$ to 0.4 is good enough.	
	When we continue to increase the $\lambda$, the semantics information from masked semantics modeling will be largely ignored, which degrades the performance.

	\subsubsection{Impact of Semantic-Adaptive Margin Loss}
	To demonstrate the effectiveness of our proposed semantic-adaptive margin loss~(SAM), we compare SAM with three different methods as the training objective: 1)~MMD~\cite{gretton2012kernel}: optimizes the distance between video and language graphs by maximum mean discrepancy; 2)~CL~\cite{oord2018representation}: optimizes the distance between video and language graphs by contrastive learning; 3)~GAN~\cite{ajakan2014domain}: measures the domain discrepancy using a discriminator network. We report the performance comparison in Table~\ref{t7}. We observe that directly minimizing the discrepancy between two graphs using the varied methods is not suitable for the partial matching problem between the video and the query. In contrast, our semantic-adaptive margin loss adaptively adjusts the soft margin according to the semantic overlap degree between two graphs, thus achieving the best performance.

	\begin{table}[!t]
		\caption{Quantitative results about the evaluation of the temporal grounding models' word order sensitivity.}
		\label{t8}
		\resizebox{\linewidth}{!}{
			\centering
			\begin{tabular}{l ccc ccc}
				\hline
				\multirow{2}*{Method}
				&\multicolumn{3}{c}{Charades-CG} 
				&\multicolumn{3}{c}{ActivityNet-CG}
				\\
				
				\cmidrule(lr){2-4} \cmidrule(lr){5-7} 
				&Trivial &Comp &\multicolumn{1}{c}{Word}   &Trivial &Comp &Word  \\
				\hline    
				2D-TAN       &0.41   &0.52   &0.43   &0.29   &0.30  &0.41 \\
				LGI              &0.28   &0.23   &0.16   &0.31   &0.22  &0.19 \\
				VLSNet       &0.07   &0.24   &0.10   &0.24   &0.31  &0.48 \\
				\hline
				VISA + ASSL &25.52 &32.03 &35.30  &24.51 &29.84  &34.21 \\
				VISA            &24.14  &29.80     &33.97   &22.09  &27.60   &31.89  \\
				\quad w/o SCL  &19.64   &24.31   &29.72   &18.07    &24.64 &28.73  \\
				\quad w/o VCC  &21.32   &26.73   &30.88   &20.15   &25.46  &29.79  \\
				\hline
			\end{tabular}
		}
	\end{table}

	\begin{figure*}[!t]
		\centering
		\includegraphics[width=0.85\linewidth]{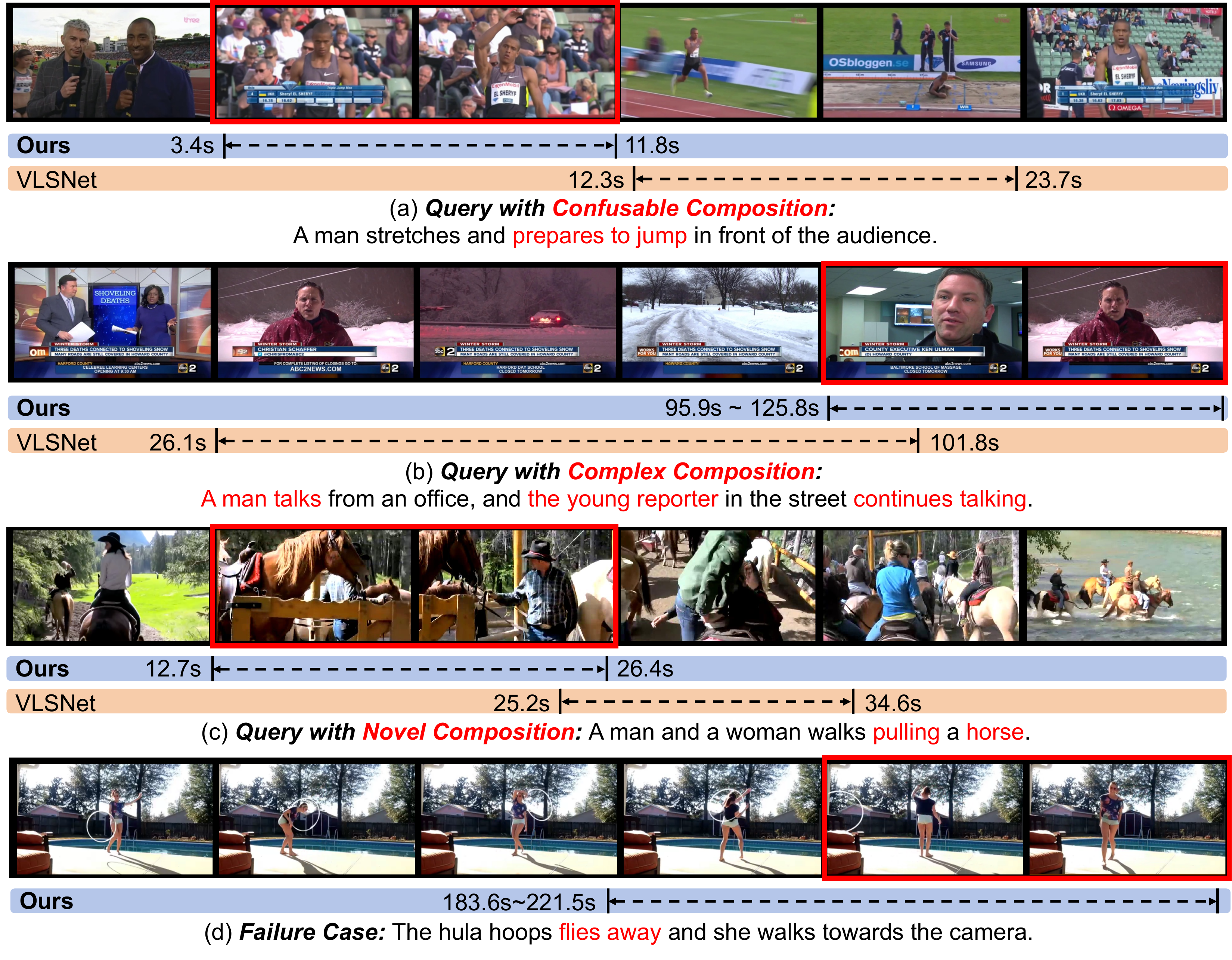}
		\caption{Qualitative examples of our model (VISA + ASSL) and VLSNet. The red boxes represent the ground-truth.}
		\label{case}
	\end{figure*}

	\subsection{In-Depth Analysis on Compositional Generalization} 
	
	\subsubsection{Results on Different Composition Types}
	To gain further insight, we examine the detailed performance (R@1, IoU=0.5) on different types of compositions. As shown in Table~\ref{t6}, our model outperforms the baselines across all composition types, and the newly proposed ASSL approach makes consistent improvement on the VISA model over all composition types. The baselines exhibit consistent performances across different composition types. We speculate it is because they are insensitive to the fine-grained semantics and simply align global query representations to the corresponding visual patterns~(mainly actions). As the Verb-Noun type changes the action appearance greater, the performance of baselines is relatively low. Also, we observe that generalizing to ``Verb-Noun'' compositions is the most difficult for our approach, as it requires the model to accurately identify the corresponding action and objects in videos and jointly reason over them to infer the semantics of the novel composition.
	
	\subsubsection{Word Order Sensitivity for Compositional Reasoning}
	
	Perceiving the syntax of a sentence is taken to be a prerequisite for understanding the compositional semantics of a sentence~\cite{sinha2020unnatural, kratzer1998semantics}. Intuitively, if we permute the word order of a sentence, its compositional semantics might be greatly distorted and thus the original ground-truth temporal boundaries might not be suitable for the shuffled sentence (in other words, the models should not predict original temporal boundaries). Since the compositionality of human language lies in the syntactic structure of sentences, we further explore whether the models are sensitive to the word order, which serves as a proxy task to quantify the syntactic understanding of current models. 
	
	\noindent
	\textbf{Evaluation.} We randomly shuffle queries in advance and then use the shuffled queries to train and evaluate the models. We define the sensitivity metric as the relative performance degradation of the shuffled version on R@1, IoU=0.5. The higher value indicates a higher sensitivity. Formally, let $R_{orig}$ and $R_{shuff}$ denote the performance on the original and shuffled queries, respectively, and the relative performance degradation can be defined as: $|R_{orig} - R_{shuff}|\ /\          R_{orig}$. We shuffle queries with different random seeds for three times and report the averaged results.
		
	\noindent
	\textbf{Results.} According to Table~\ref{t8}, we surprisingly find that previous SOTA models are insensitive to the word order. This indicates that the performance gains of existing temporal grounding models mainly come from exploring superficial correlations in the datasets instead of understanding compositional semantics. Thus, these models fail to ground novel compositional semantics of language queries into video context. In contrast, our models with varied components are sensitive to the compositional structure of sentences. Distorting the compositional structure of query sentences seriously affects the performance of our models. Particularly, we observe the highest sensitivity on novel-word splits, indicating that the compositional structure is important for inferring the semantics of novel words. Moreover, comparing VISA + ASSL with VISA, we find that the proposed ASSL approach increases the sensitivity as it requires to infer the structured semantics from video and language. In the end, we observe that the proposed SCL and VCC promote our model to capture the compositional structure of sentences in a mutually rewarding way.

	\begin{table}[!t]
		\caption{Comparison of model complexity with regard to model size, FLOPs, and throughput.}
		\label{t10}
		\centering
		\begin{threeparttable}
			\begin{tabular}{ l ccc}
				\hline
				Method          &Parameters  &FLOPs  &Throughput \\				
				\hline    
				2D-TAN         &52.9M     &942.7G            &200.1 video/s \\
				LGI                 &19.3M        &66.3G              &440.1 video/s\\
				\hline
				VISA              &15.6M     &37.9G     &456.8 video/s \\
				VISA + ASSL   &15.7M   &38.4G   &455.2 video/s    \\
				\hline
			\end{tabular}
		\end{threeparttable}
	\end{table}

	\subsection{Model Efficiency Comparison}
	We report the complexity analysis in Table~\ref{t10}. The experiments are run with an NVIDIA 2080Ti GPU and a batch size of 32. It can be observed that our models achieve larger throughput with the smallest model size. The main reasons are two-folds: \textbf{i)} our hierarchical semantic graph avoids dense computation between low-level frame features and word features because of its multi-level structured modeling of videos and sentences; \textbf{ii)} our variational cross-graph correspondence reasoning directly infers the target moment instead of generating masses of proposals with various sliding windows in the video.

	\begin{figure*}[!t]
		\centering
		\includegraphics[width=0.85\linewidth]{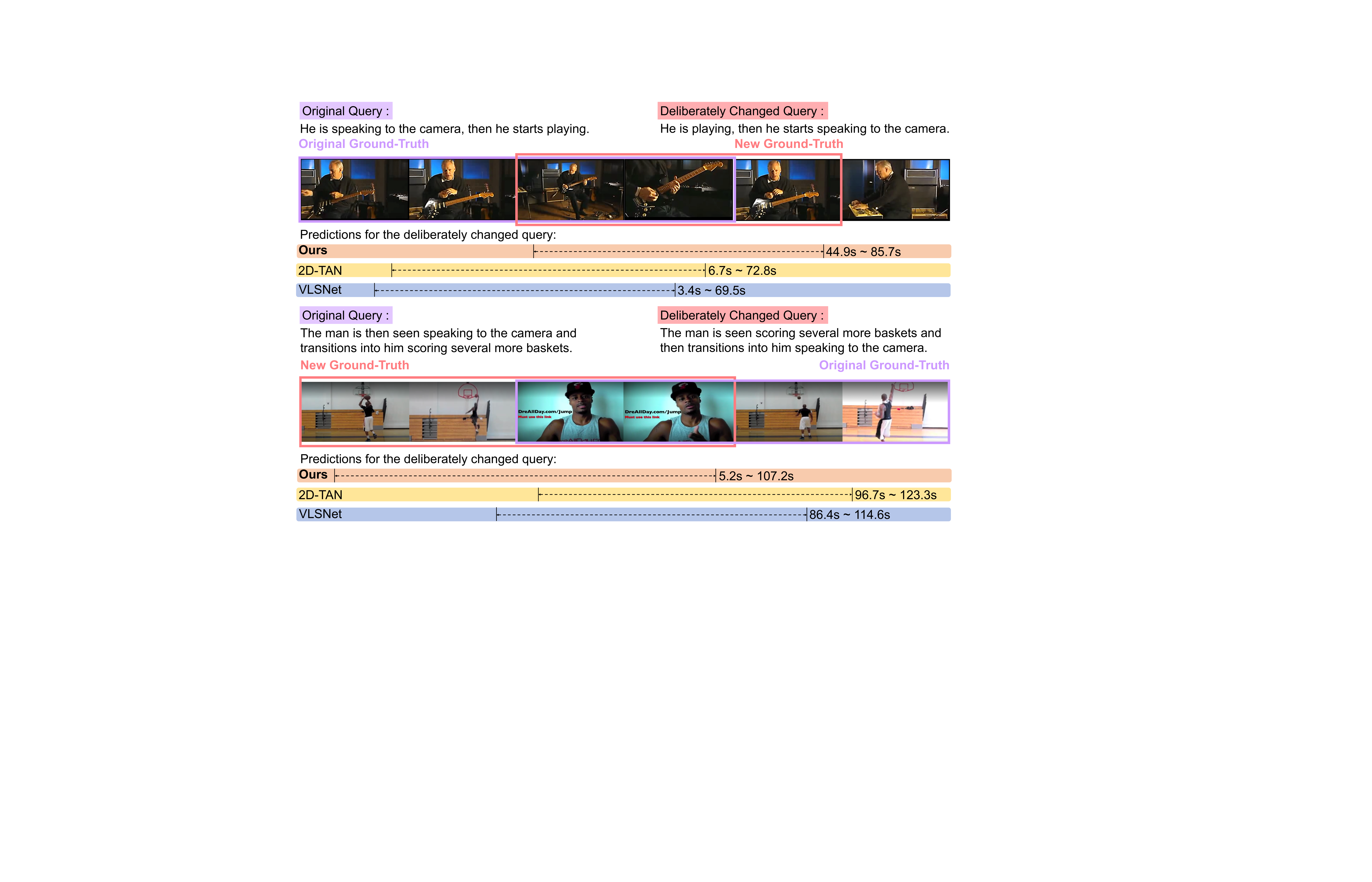}
		\caption{Qualitative examples on specifically shuffled queries. The SOTA methods fail to capture the semantics change of the deliberately changed query, whereas our model (VISA + ASSL) successfully locates to the new temporal boundary.}
		
		\label{shuffle}
	\end{figure*}

	\subsection{Qualitative Analysis}
		
	\subsubsection{Qualitative Examples}
	Figure \ref{case} (a)-(c) visualizes three successful qualitative examples, which indicate the importance of compositionality. In the first case, the baseline fails to understand the composition meaning of ``prepares to jump '', so it mistakenly localizes to the ``jump'' segment. In contrast, our VISA successfully captures the compositional meanings. The second case contains complex compositions, which describe two events. Without inferring their temporal relationship from the composition structure, the baseline localizes a wrong segment, even though it also contains the two individual events (\textsl{i.e., ``a man talk''} and \textsl{``the reporter in the street talks''}). Conversely, our VISA understands the correct temporal order of these two events. The third case shows that our VISA successfully generalizes to novel composition. While \textsl{``pulling''}~(\textsl{e.g., pulling rope}) and \textsl{``horse''}~(\textsl{e.g., lead horse}) are both observed in the training split, the baseline suffers from generalizing to this novel composition.

	\subsubsection{Sensitivity on Specific Shuffling}
	We manually select some query sentences and change their word order in some specific ways, such that the changed query can still semantically correspond to other segments in their original videos. As shown in Figure \ref{shuffle}, we annotate the changed query with a new ground-truth~(red box) and use the changed query to test models. Interestingly, the predictions of SOTA methods have higher IoU with the original temporal boundary, though the semantics of the sentence has been deliberately modified. This phenomenon further indicates that previous SOTA methods are insensitive to the word order and thus fail to capture the semantics change of the deliberately changed query. In contrast, our model keenly captures the semantics change and locates to the new temporal boundary.
	
	\begin{figure}[!t]
		\centering
		\includegraphics[width=\linewidth]{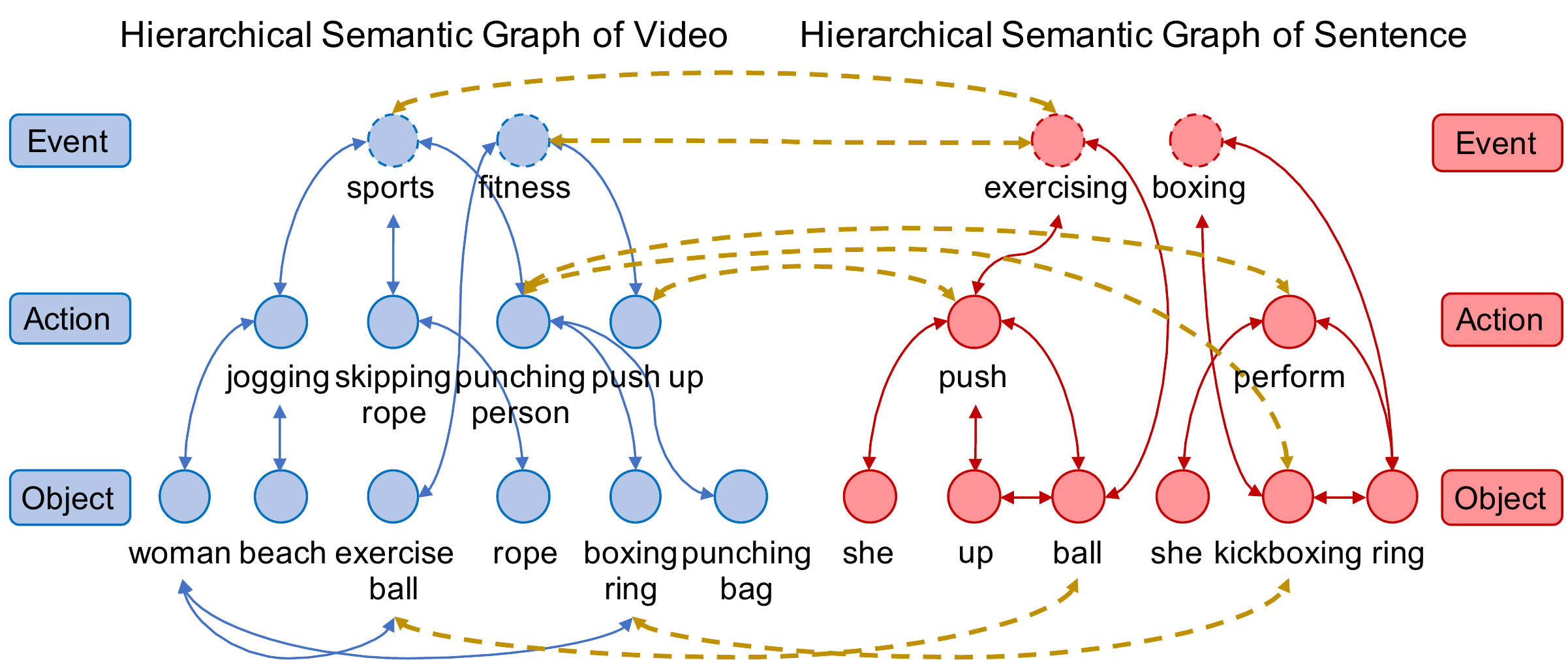}
		\caption{Visualization of the learned hierarchical semantic graph.
		}
		\label{vis}
	\end{figure}	
	
	\subsubsection{Error Analysis}
	In this section, we further analyze the negative samples and showcase a typical failure case in Figure~\ref{case}~(d). As shown in Figure~\ref{case}~(d), our model cannot discriminate subtle semantics of adverbs~(\textsl{i.e., flies away}) and therefore fails to identify the accurate start point of  ``the hula loops flies away''. Similarly, we find it is hard for our model to identify ``pull slightly''  and ``slice again'', of which the semantics is fine-grained. We expect future research to utilize the new benchmarks to make progress on fine-grained semantics grounding, thus achieving compositional generalization.

	\subsubsection{Visualizing Learned Graph}
	In Figure \ref{vis}, we present the learned hierarchical semantic graph. We visualize some key nodes and the edges with high weights. The yellow dotted lines represent the cross-graph semantic correspondence. If the semantic correspondence score between two nodes is greater than a specific threshold, we connect them with a yellow dotted line. We represent the event nodes by their most related semantics according to their attended nodes. Our VISA successfully aligns the visual semantics ``punching person'' and ``boxing ring'' to the linguistic words ``perform kickboxing''.  Also, our VISA can connect ``push up'' and ``exercise ball'' to the words ``push up (with) ball''.

	\section{Conclusion}
	In this paper,  we introduce a new task, compositional temporal grounding, as well as two newly organized datasets to systematically benchmark the compositional generalizability of temporal grounding models. We conduct in-depth analyses on SOTA methods and find they struggle to achieve compositional generalization. We then introduce a novel variational cross-graph reasoning and adaptive structured semantics learning framework to achieve compositional generalizable temporal grounding models. Specifically, we first propose a variational cross-graph reasoning framework that learns fine-grained semantic correspondence between video and language by explicitly decomposing them into hierarchical semantic graphs. Second, we present a novel adaptive structured semantics learning approach to encourage the hierarchical semantic graph to learn structure-informed and domain-generalizable graph representations. Experiments illustrate the significant effectiveness and efficiency of our model. Particularly, our model significantly surpasses the SOTA methods on the novel-composition and novel-word splits of Charades-CG and ActivityNet-CG datasets while maintaining the lower complexity by explicit graph reasoning.

	
	%



	{
		\small
		\bibliographystyle{ieee_fullname}
		\bibliography{pami}
	}

\end{document}